\documentclass[letterpaper]{article} % DO NOT CHANGE THIS
\usepackage{aaai25}  % DO NOT CHANGE THIS
\usepackage{times}  % DO NOT CHANGE THIS
\usepackage{helvet}  % DO NOT CHANGE THIS
\usepackage{courier}  % DO NOT CHANGE THIS
\usepackage[hyphens]{url}  % DO NOT CHANGE THIS
\usepackage{graphicx} % DO NOT CHANGE THIS
\usepackage{natbib}  % DO NOT CHANGE THIS AND DO NOT ADD ANY OPTIONS TO IT
\usepackage{caption} % DO NOT CHANGE THIS AND DO NOT ADD ANY OPTIONS TO IT
\usepackage{algorithm}
\usepackage{algorithmic}
\usepackage{comment}

\usepackage{newfloat}
\usepackage{listings}
\DeclareCaptionStyle{ruled}{labelfont=normalfont,labelsep=colon,strut=off} % DO NOT CHANGE THIS
\floatstyle{ruled}
\newfloat{listing}{tb}{lst}{}
\floatname{listing}{Listing}
\title{Safe Interval RRT$^*$ for Scalable Multi-Robot Path Planning in Continuous Space}
\author {
    Joonyeol Sim\textsuperscript{\rm 1},
    Joonkyung Kim\textsuperscript{\rm 1},
    Changjoo Nam\textsuperscript{\rm 1}
}
\affiliations {
    \textsuperscript{\rm 1}Sogang University, Republic of Korea\\
    jysim@u.sogang.ac.kr, 
    jkkim@u.sogang.ac.kr, 
    cjnam@sogang.ac.kr
}

\usepackage{bibentry}
\usepackage{xspace}
\usepackage{xcolor}
\usepackage{amsmath,amsfonts}
\usepackage{amsthm}
\usepackage{multirow}
\usepackage{multicol}

\newtheorem{theorem}{Theorem}

\newcommand{\sirrt}{SI-RRT$^*$\xspace}
\newcommand{\strrt}{ST-RRT$^*$\xspace}
\newcommand{\strrtpp}{ST-RRT$^*$\hspace{-2pt}-PP\xspace}
\newcommand{\sicbs}{SI-CCBS\xspace}
\newcommand{\sipp}{SI-CPP\xspace}

\providecommand{\vt}[1]{\boldsymbol#1}

\begin{document}

\maketitle

\begin{abstract}
In this paper, we consider the problem of Multi-Robot Path Planning (MRPP) in continuous space. The difficulty of the problem arises from the extremely large search space caused by the combinatorial nature of the problem and the continuous state space. We propose a two-level approach where the low level is a sampling-based planner \textit{Safe Interval RRT$^*$ (SI-RRT$^*$)} that finds a collision-free trajectory for individual robots. The high level can use any method that can resolve inter-robot conflicts where we employ two representative methods that are Prioritized Planning (\sipp) and Conflict Based Search (\sicbs). Experimental results show that \sirrt can quickly find a high-quality solution with a few samples. \sipp exhibits improved scalability while \sicbs produces higher-quality solutions compared to the state-of-the-art planners for continuous space. %Compared to the most scalable existing algorithm, \sipp achieves a success rate that is up to 94\% higher with 100 robots while maintaining solution quality (i.e., \textit{flowtime}, the sum of travel times of all robots) without significant compromise. \sipp also decreases the makespan up to 45\%. \sicbs decreases the flowtime by 9\% compared to the competitor, albeit exhibiting a 14\% lower success rate.
\end{abstract}


\section{Introduction} 
\label{sec:intro}
\begin{figure}[t]
\centering
\captionsetup{skip=0pt}
\includegraphics[width=0.99\linewidth]{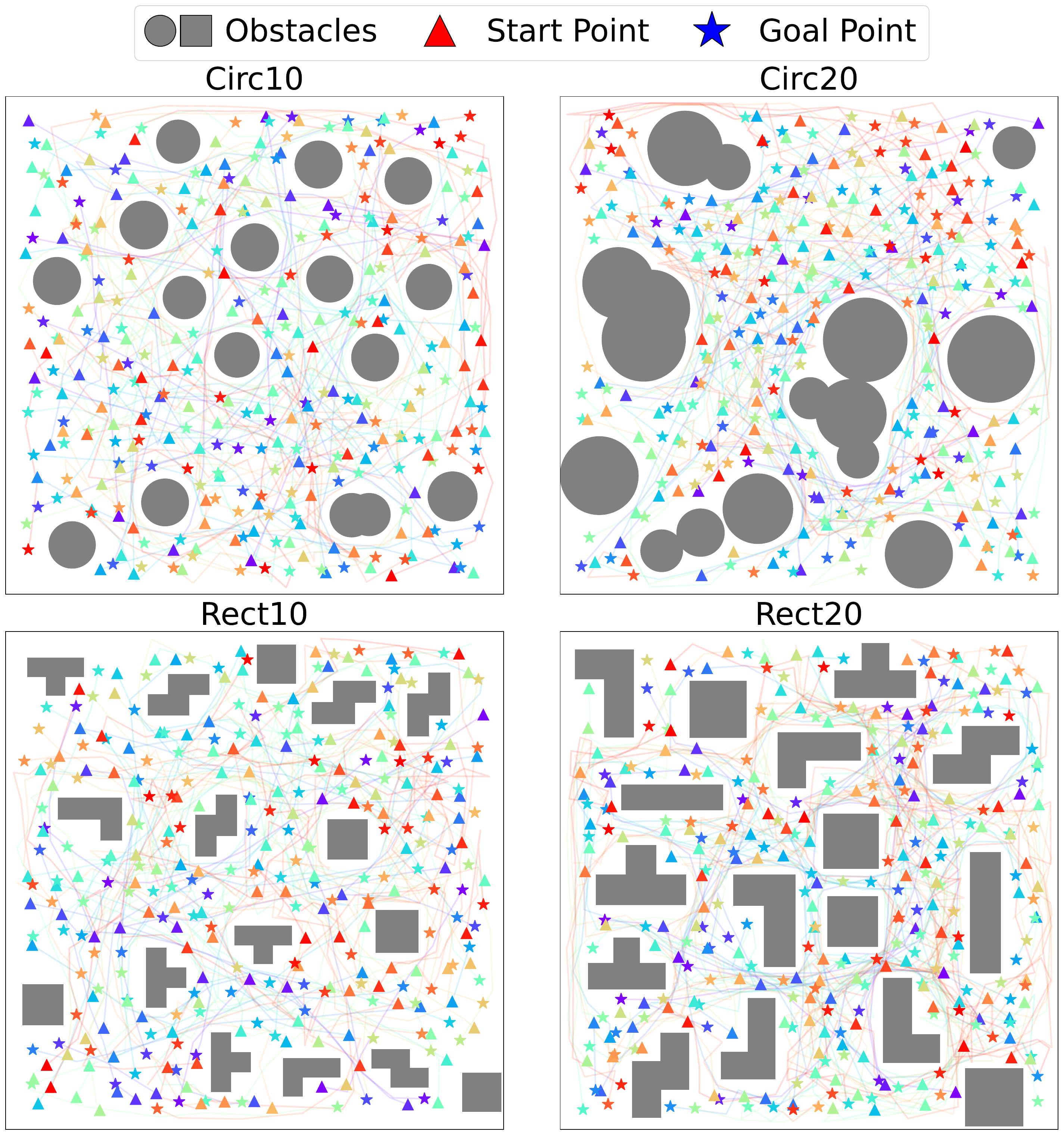}
\caption{Challenging instances with 160 robots where robots often experience conflicts.  These are also test environments \textsf{\small Circ10}, \textsf{\small Circ20}, \textsf{\small Rect10}, and \textsf{\small Rect20} used in our experiments. 
}\vspace{-20pt}
\label{fig:env}
\end{figure}

% Definition of MRPP
Multi-Robot Path Planning (MRPP) is the problem of finding collision-free paths for multiple robots while optimizing a given objective, such as the total path length (i.e., sum of costs) or the time taken to complete all tasks (i.e., makespan)~\cite{mapfdefinition}, even in congested environments as shown in Fig.~\ref{fig:env}.
% Is it a proper technique?
While the problem has significant real-world applications like warehouse automation~\cite{warehouse} and search and rescue~\cite{searchandrescure}, finding optimal solutions is computationally challenging, as proven to be NP-hard~\cite{nphard}.
% What is the previous approach for MRPP?

% Simple review of a category of an existing approach for multi-agent pathfinding
% 마지막 문장을 조금 더 구체적으로 써야 discreted말고 continuous에서 풀어야 하는 문제를 직관적으로 전달할 수 있을 것 같습니다.
MRPP has been extensively studied in the context of multi-agent systems. Existing methods can be broadly classified into  \textit{coupled}, \textit{decoupled}, and \textit{dynamically-coupled}~\cite{mstar}. The coupled approach (e.g., \cite{od}, \cite{epeastar}) does not scale with the number of robots, although they guarantee completeness and optimality. The decoupled approach (e.g., \cite{ppvariants}, \cite{hcastar}) reduces the search space for efficiency but does not provide theoretical guarantees. Dynamically-coupled methods (e.g., \cite{cbs}, \cite{mstar}) combine the advantages of the former two by coupling robots only whenever necessary. However, these methods often struggle to scale and require discretizing continuous environments, leading to suboptimal or infeasible solutions when applied to real-world robotic systems.
%This highlights the need for a fast continuous-space MRPP algorithm.

% Present our goal
Our goal is to develop an efficient method for MRPP that can handle more than 100 robots in continuous space.
We propose a two-level approach consisting of a low-level single-robot planner and a high-level conflict-resolving method. For the low level, we propose \textit{Safe Interval RRT$^*$} (\sirrt), which borrows the concept of safe time intervals from Safe Interval Path Planning (SIPP)~\cite{phillips2011sipp}. While SIPP requires discretization of the environment, \sirrt allows a robot to navigate in continuous space without discrete time steps. The high level can use any conflict resolution method, such as Prioritized Planning (PP) and Conflict Based Search (CBS). We develop \textit{Safe Interval Continuous-space PP} (SI-CPP) and \textit{Safe Interval Continuous-space CBS} (SI-CCBS), which have their own advantages in computational efficiency and solution quality, respectively.
% Possibility of applying kinodynamics
% Furthermore, our suggested approach has the potential to include kinodynamic constraints by utilizing a kinodynamic local planner between two states, making it more suitable for real-world scenarios.

% main contributions
The main contributions of the proposed method are (i) the improved scalability as \sipp can handle more than 160 robots in continuous space even in densely cluttered environments and (ii) theoretical guarantees for the probabilistic completeness and asymptotic optimality of \sirrt. (iii) In addition, we extend the proposed methods to consider the kinodynamic constraints of robots for real-world applications. We also provide extensive evaluations comparing our methods with state-of-the-art approaches such as SSSP~\cite{sssp}, ST-RRT$^*$-PP~\cite{strrtstar}, and Graph Transformer~\cite{yu2023accelerating} in terms of computational efficiency and solution quality.

\vspace{-5pt}
\section{Related Work}
\label{sec:related}
% New커멘트 : 생략해도 될 것 같고 대신 섹션 마지막에 정리하는 단락을 추가했습니다.
% introduction
%As our goal is to develop an MRPP method in continuous space, we review several recent works on MRPP and multi-agent pathfinding (MAPF) methods in continuous space.

% SIPP
SIPP~\cite{phillips2011sipp} is an efficient single-robot path planner that can avoid dynamic obstacles. A \textit{safe interval} is defined as a time period for a specific configuration during which no collision occurs. Since the number of safe intervals is significantly smaller than the number of time steps, SIPP effectively reduces the dimensionality of the problem. Nevertheless, as SIPP is designed for discrete space, it has limitations in handling robots that operate in continuous spatial and temporal spaces. 

% ST-RRT*
ST-RRT$^*$~\cite{strrtstar} is a bidirectional planner working in a space-time configuration space, which consists of Cartesian coordinates and a time dimension. PP~\cite{orthey2020multilevel} is used to resolve conflicts, where robots with lower priority modify their individual paths to avoid higher-priority robots. Since the search space has a time dimension, the robots can exploit temporality for more effective collision avoidance. Although PP does not guarantee deadlock-free paths, \strrtpp exhibits remarkable scalability and high-quality solutions. 

% SSSP
Simultaneous Sampling and Search Planning (SSSP)~\cite{sssp} is a unified approach that does not separate the roadmap construction and conflict resolution. SSSP not only finds solutions quickly but also guarantees probabilistic completeness. However, the solution of SSSP is \textit{sequential}, meaning that only one robot can move each time. A postprocessing method is necessary to make multiple robots move simultaneously, which requires additional computations. The probabilistic completeness would not hold for this post-processed solution. %\js{we should check the last sentence}

% GT
In~\cite{yu2023accelerating}, a Graph Transformer (GT) is used as a heuristic function to accelerate CBS in non-grid settings, aiming for completeness and bounded-suboptimality. A contrastive loss training objective is introduced for learning a heuristic that ranks search nodes, demonstrating the generalizability of the method with promising results in accelerating CBS. While the approach of learning the heuristic function is novel, the success rate of the search decreases significantly as the number of robots increases. 

% New커멘트 : 정리/conclusion
Among the existing methods, \strrtpp is the most scalable and practical for MRPP or multi-agent pathfinding (MAPF) in continuous space. We aim to develop a more efficient algorithm with improved solution quality and scalability, particularly for large numbers of robots in complex environments, to push the limit of the current state-of-the-art methods.

% free space가 중복되어서 free workspace로 변경했습니다.
\vspace{-5pt}
\section{Problem Definition}
\label{sec:prob}
% Workspace
We consider the MRPP problem of finding collision-free trajectories for $k$ robots navigating from their start locations to respective goal locations in a \textit{workspace} $\mathcal{W} \subset \mathbb{R}^d$ where $d \in \{2, 3\}$. 
% Obstacles
A point $\vt{w} \in \mathcal{W}$ can be occupied by an obstacle where the sets of static and dynamic obstacles are $\mathcal{O}_S$ and $\mathcal{O}_D$, respectively. We assume that $\mathcal{O}_S$ and $\mathcal{O}_D$ are  known.
% Free space
The \textit{free workspace} is $\mathcal{W}^{f} = \mathcal{W} \setminus \mathcal{O}$ where $\mathcal{O} = \mathcal{O}_S \cup \mathcal{O}_D$. 
% Configuration Space
The \textit{configuration space} (C-space) $\mathcal{C}^i \subseteq \mathcal{W}$ is the set of configurations of robot $i$. 
% The shape of the robot
For a configuration $\vt{q} \in \mathcal{C}^i$, the set of points occupied by $i$ at $\vt{q}$ is denoted as $h^i(\vt{q}) \subset \mathcal{W}$, where $h^i(\vt{q})$ depends on the shape of robot $i$. The free C-space (\textit{free space}) is $ \mathcal{C}^i_\text{free} = \{ \vt{q} \in \mathcal{C}^i \mid h^i(\vt{q}) \cap \mathcal{O} = \emptyset \}$.
% Trajectory
A trajectory of robot $i$ is a continuous function $\pi^i : [0, t^i_\text{final}] \rightarrow \mathcal{C}^i_\text{free}$ where $t^i_\text{final}$ means the last time of the trajectory of $i$. Each configuration $\vt{q}$ in a trajectory is given by $\pi^i(t)$ for $t \in [0, t^i_\text{final}]$. 
% execution time
% Given any pair of configurations in $\pi^i$, the velocity (so the time) to move between them is determined by the kinodynamic constraints of $i$.

\begin{figure}[t]
\captionsetup{skip=0pt}
\centering
\includegraphics[width=0.75\columnwidth]{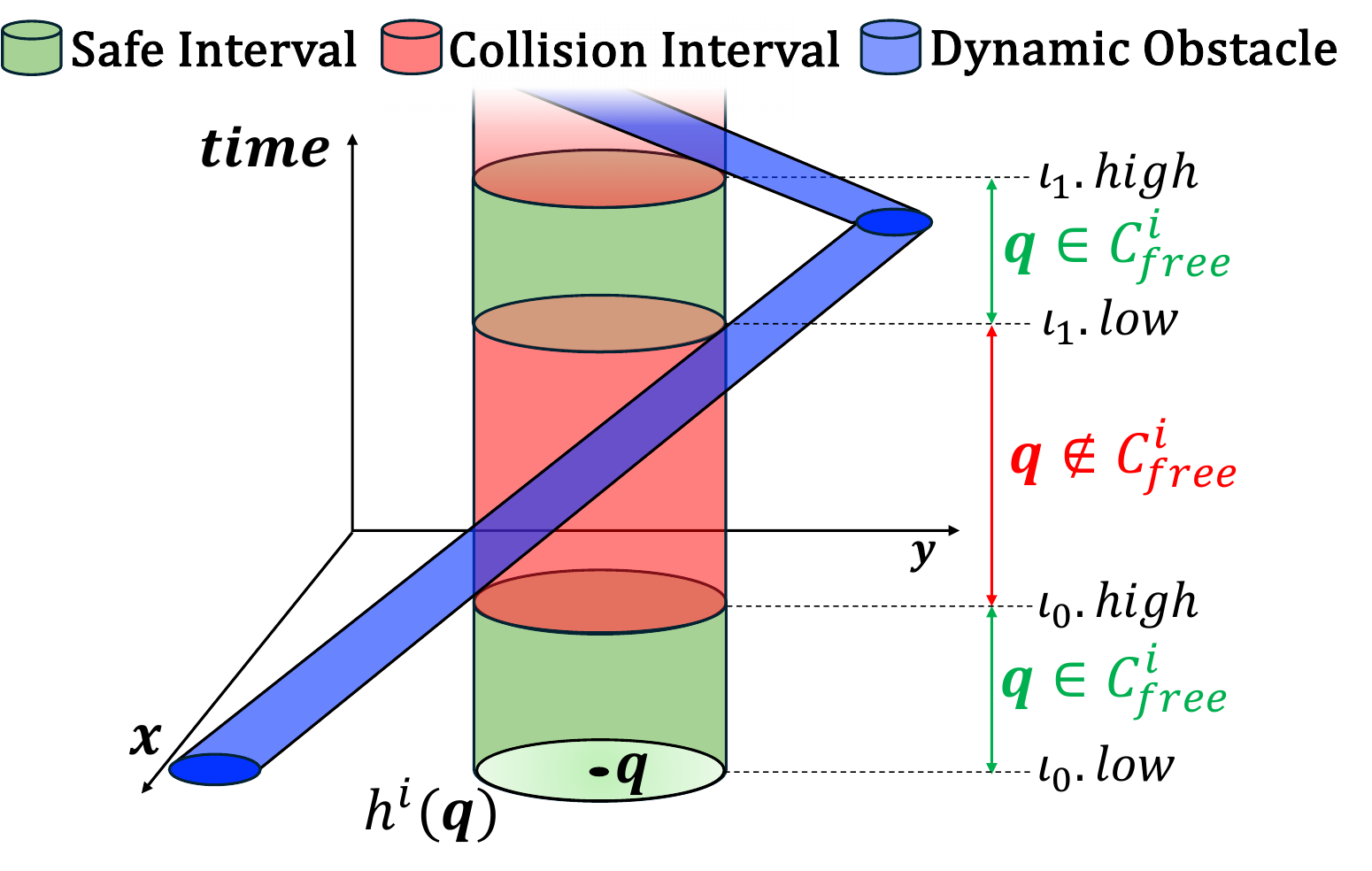}
\captionsetup{skip=0pt}
\caption{An illustration of the safe interval at $\vt{q}$. The interval is defined for $h^i(\vt{q})$ to consider the size of robot $i$ at $\vt{q}$. The robot can stay at $\vt{q}$ without a collision during the safe interval (green). The shapes of the robot and dynamic obstacles, as well as the trajectory of the obstacle, can be arbitrary.}\vspace{-12pt}
\label{fig:si}
\end{figure}

% Define Safe Interval
% With a slight abuse of notation이 너무 강한 느낌이어서 For convenience가 어떨까 싶습니다.
A \textit{safe interval} (SI) of a configuration $\vt{q}$ is a continuous time period $\iota = [low, high)$ during which the robot can remain at $\vt{q}$ without collisions, meaning $\vt{q}$ resides in $\mathcal{C}^i_\text{free}$ throughout $\iota$. For convenience, we use $\iota.low$ and $\iota.high$ to denote the start and end times of this interval. Within this SI, the robot cannot arrive at $\vt{q}$ earlier than $\iota.low$ and must leave $\vt{q}$ no later than $\iota.high$ to avoid collisions. 
% Define Collision Interval
Conversely, a \textit{collision interval} (CI) for $\vt{q}$ is a time period during which the robot at $\vt{q}$ would collide with at least one obstacle (i.e., a period when $\vt{q} \notin \mathcal{C}^i_\text{free}$). 
% Alternating
An arbitrary time period for any $\vt{q}$ can be partitioned into alternating SIs and CIs depending on the trajectories of dynamic obstacles. 
% Example of safe interval
In Fig.~\ref{fig:si}, the points occupied by robot $i$ (i.e., $h^i(\vt{q})$) are depicted as a circle, which can represent any arbitrary shape. Green zones denote SIs where $h^i(\vt{q}) \cap \mathcal{O} = \emptyset$ and red zones indicate CIs when the dynamic obstacle (blue) intersects $h^i(\vt{q})$.

% How to deal with too many safe intervals
Unlike the discrete space with finite states, a continuous space has an unbounded number of configurations. Since we cannot enumerate all SIs, we construct a \textit{safe interval map} $\mathcal{M}$ to store SIs whenever necessary. An SI at $\vt{q}$ is added to $\mathcal{M}$ only if $\vt{q}$ is considered during the planning process. 

% Define roadmap
For robot $i$, we define a roadmap (or a tree) $G^i = (\mathcal{V}, \mathcal{E})$ which is a connected topological graph within a $(d+1)$ dimensional space where the last one dimension represents time. A vertex $v \in \mathcal{V}$ in $G^i$ corresponds to a specific configuration $\vt{q}$ and its earliest arrival time $v.t_\text{low}$, indicating that the robot cannot reach this configuration earlier than $t_\text{low}$. An edge $e \in \mathcal{E}$ connects a pair of vertices in $\mathcal{V}$ if a feasible trajectory exists between their respective configurations, given the SIs of these configurations. 

% Define Conflict and Constraints
A \textit{conflict} $\langle i, j, h^i(\pi^i(t)), h^j(\pi^j(t)) \rangle$ is defined for robots $i$ and $j$ for $t \in [t_s, t_e]$ where $h^i(\pi^i(t))$ represents the area occupied by $i$ at $t$ while $i$ moves along $\pi^i$. Robots $i$ and $j$ collide within $[t_s, t_e]$ so $h^i(\pi^i(t)) \cap h^j(\pi^j(t)) \neq \emptyset$ for $t \in [t_s, t_e]$. A \textit{constraint} for robot $i$ is $\langle i, h^j(\pi^j(t)), t \rangle$ where $t \in [t_s, t_e]$ indicating that robot $i$ must avoid $h^j(\pi^j(t))$ during $[t_s, t_e]$. In other words, robot $i$ must generate a trajectory while satisfying $h^i(\pi^i(t)) \cap h^j(\pi^j(t)) = \emptyset$ for $t \in [t_s, t_e]$.

For a single robot $i$, we aim to find a collision-free trajectory $\pi^i$ from the start configuration $\vt{q}^i_\text{start}$ to the goal configuration $\vt{q}^i_\text{goal}$ with the minimum arrival time $t^i_\text{final}$. 
For multiple robots, we aim to find collision-free trajectories for all $k$ robots while minimizing the \textit{flowtime} $\kappa$, which is the sum of individual arrival times of all robots at their goal configurations. We also aim to have a small \textit{makespan} $\tau$, which is the elapsed time until all robots reach their goals. They are defined as $\kappa = \sum_{i=1}^{k} t^i_\text{final}$ and $\tau = \max \{t^i_\text{final}| t^i_\text{final} \in \mathbb{R}, 1 \leq i \leq k\}$, respectively. 
The collision-free trajectories satisfy $h^i(\pi^i(t)) \cap h^j(\pi^j(t)) = \emptyset \ \forall i, j, i \neq j, \forall t \in [0, {t_\text{final}}]$ where ${t_\text{final}} = \max \{t^1_\text{final}, \cdots, t^k_\text{final}\}$.

\vspace{-5pt}
\section{Single-Robot Planner: Safe Interval RRT$^*$}
\label{sec:single}
% Overview of SIRRT
We begin with proposing a sampling-based planner \sirrt for individual robots, which does not need to simplify the environment by discretizing the space and time. Also, fewer samples are needed to find high-quality trajectories compared to existing methods. For convenience, we temporarily omit the robot index (e.g., $i$) throughout this section.

% improving feature of sirrt
\sirrt expands $G$ up to $iteration$ times to find a collision-free trajectory $\pi(t)$ for $t \in [t_\text{start}, t_\text{goal}]$. Once a solution is found, it continues improving it with more samples within the $iteration$ limit.
Here, $t_\text{start}$ and $t_\text{goal}$ represent the times at $\vt{q}_\text{start}$ and $\vt{q}_\text{goal}$ of the robot, respectively. These configurations belong to vertices $v_\text{start}$ and $v_\text{goal}$ of $G$. 

\vspace{-5pt}
\subsection{Functions for tree expansion}
\label{sec:expfunc}
Functions \textsc{Sampling} and \textsc{Steer} are used to grow $G$. 
% Sampling
\textsc{Sampling} randomly selects $\vt{q}_{\text{rand}}$ from $\mathcal{C}$ with a probability of $1-\lambda$ or the goal configuration $\vt{q}_\text{goal}$ with a probability of $\lambda$ where $\lambda \in [0, 1]$.
% Steer
\textsc{Steer} attempts to generate a new configuration $\vt{q}_{\text{new}}$. It identifies $v_\text{near}$ in $G$ whose configuration $v_\text{near}.\vt{q}$ is the nearest to $\vt{q}_{\text{rand}}$. If the distance between these two configurations is within a threshold $d_\text{max}$, $\vt{q}_{\text{new}}$ is set to $\vt{q}_{\text{rand}}$. Otherwise, $\vt{q}_{\text{new}}$ is placed along the line segment connecting $v_\text{near}.\vt{q}$ and $\vt{q}_{\text{rand}}$, at a distance of $d_\text{max}$ from $v_\text{near}.\vt{q}$.
\textsc{Steer} performs collision checking. If the trajectory from $v_\text{near}.\vt{q}$ to $\vt{q}_\text{new}$ results in a collision with any obstacle in $\mathcal{O}$, it returns \text{NULL}. Otherwise, it returns $\vt{q}_{\text{new}}$.

\begin{figure}[t]
\captionsetup{skip=0pt}
\centering
\includegraphics[width=\columnwidth]{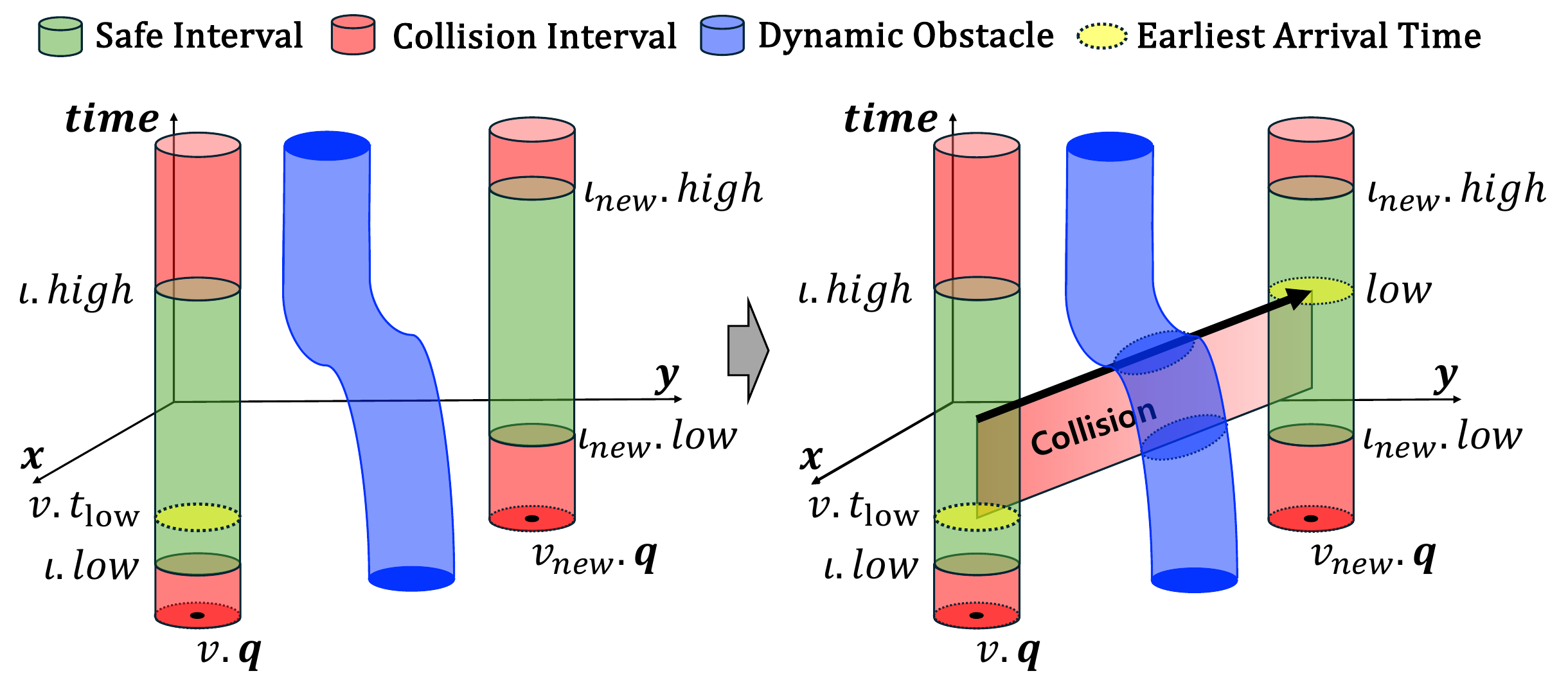}
\caption{An illustration showing how $low$ is determined in \textsc{ChooseParent}. (Left) The robot can stay $v$ from $v.t_{\text{low}}$ until $\iota.high$ without collisions. The SI of $v_\text{new}.\vt{q}$ is $[\iota_\text{new}.low, \iota_\text{new}.high)$. (Right) The bold arrow represents a trajectory to move from $v.\vt{q}$ to $v_\text{new}.\vt{q}$ as early as possible without colliding with the blue dynamic obstacle. Thus, $low$ is the earliest arrival time at $v_\text{new}.\vt{q}$ along the trajectory. 
}\vspace{-12pt}
\label{fig:low}
\end{figure}

\vspace{-5pt}
\subsection{Functions for tree optimization}
\label{sec:optfunc}
Functions \textsc{GetNeighbor}, \textsc{ChooseParent}, and \textsc{Rewire} are used to optimize $G$.\footnote{Owing to the space limit, we provide pseudocodes of these important subroutines \textsc{ChooseParent} and \textsc{Rewire} as supplementary materials.}
% Get Neighbor
\textsc{GetNeighbor} returns the set of nearby vertices $\mathcal{V}_\text{neighbor}$ which are within $d_\text{max}$ from $\vt{q}_{\text{new}}$. It performs collision checking to ensure that there is no collision between any pair of $\vt{q}_{\text{new}}$ and a vertex in $\mathcal{V}_\text{neighbor}$.
% Overview of Choose Parent
\textsc{ChooseParent} returns a set of vertices $\mathcal{V}_\text{new}$ for $\vt{q}_{\text{new}}$ where each $v_\text{new} \in \mathcal{V}_\text{new}$ is determined for each SI of $\vt{q}_{\text{new}}$. For each $v_\text{new}$, it selects a parent from $\mathcal{V}_\text{neighbor}$ that minimizes $v_\text{new}.t_{\text{low}}$.
% Overview of Rewire
\textsc{Rewire} optimizes the tree structure of the roadmap $G$ by potentially reassigning the parent vertices of the neighbors in $\mathcal{V}_\text{neighbor}$. For each neighbor, \textsc{Rewire} checks if reassigning any $v_\text{new} \in \mathcal{V}_\text{new}$ as its new parent would result in an earlier arrival time. If so, the parent is updated accordingly.

% Detailed explanation of chooseParent
In \textsc{ChooseParent}, a new set of vertices $\mathcal{V}_\text{new}$ is initialized as empty. It iterates over all SIs associated with $\vt{q}_\text{new}$, attempting to create a new vertex for each interval. 
% New커멘트 : 아래 커멘트 아웃한 문장 어색해서 수정해봤는데 살펴봐주세요.
For each SI $\iota_\text{new}$, $v_\text{new}$ is created corresponding to $\vt{q}_\text{new}$, with $t_\text{low} = \infty$ and no parent assigned.
%For each SI $\iota_\text{new}$, $v_\text{new}$ is created with $\vt{q}_\text{new}$, $t_\text{low} = \infty$, and parent = \text{NULL}. 
It then iterates over all vertices in $\mathcal{V}_\text{neighbor}$ to determine if any of them can serve as the parent vertex for $v_\text{new}$. For each $v \in \mathcal{V}_\text{neighbor}$, it checks for a collision-free trajectory from $v$ to $v_\text{new}$ within $\iota_\text{new}$. If found, the earliest possible arrival time at $v_\text{new}$ is calculated and stored as $low$.

% Detailed explanation of how to calculate low
In the left of Fig.~\ref{fig:low}, the robot can stay at $v.\vt{q}$ from $v.t_{\text{low}}$ until $\iota.high$ without collisions. The SI of $v_\text{new}.\vt{q}$ (green) is $[\iota_\text{new}.low, \iota_\text{new}.high)$. In the right of Fig.~\ref{fig:low}, the bold arrow represents a trajectory to move from $v.\vt{q}$ to $v_\text{new}.\vt{q}$ as early as possible within their respective SIs while avoiding collisions with the dynamic obstacle (blue). Thus, the earliest arrival time at $v_\text{new}.\vt{q}$ along this trajectory is assigned to $low$.

% In case there is no low
If $\iota_\text{new}$ does not overlap with the time range during which the robot can reach $v_\text{new}.\vt{q}$ from $v$, no feasible trajectory exists between them. Even with an overlap, a feasible trajectory may not exist if the trajectory from $v$ to $v_\text{new}.\vt{q}$ inevitably collides with a dynamic obstacle, regardless of when the robot departs. In either case, $low$ cannot be determined, so the algorithm proceeds to the next vertex in $\mathcal{V}_\text{neighbor}$.
% In case there is low
If a valid $low$ is found and lower than $v_\text{new}.t_\text{low}$, both the earliest arrival time and parent vertex of $v_\text{new}$ are updated. Iterating over $\mathcal{V}_\text{neighbor}$ ensures $v_\text{new}$ to get the earliest possible arrival time in the current tree structure. After checking all neighbors for each $v_\text{new}$, if a parent has been assigned to $v_\text{new}$ (i.e., $v_\text{new}.t_\text{low} < \infty$), it is added to both $\mathcal{V}_\text{new}$ and $G$. After processing all SIs of $\vt{q}_\text{new}$, $\mathcal{V}_\text{new}$ is returned.

\textsc{Rewire} optimizes $G$ by reconnecting edges when better trajectories are possible, using $\mathcal{V}_\text{new}$ from \textsc{ChooseParent}. For each $v_\text{new} \in \mathcal{V}_\text{new}$, it evaluates whether making $v_\text{new}$ the parent of any $v \in \mathcal{V}_\text{neighbor}$ would decrease the earliest arrival time of $v$. This is done by computing the potential earliest arrival time at $v$ when reaching it via $v_\text{new}$. If a valid $low$ is found and earlier than $v.t_\text{low}$, then it updates $v.t_\text{low}$ to $low$ and reassigns the parent of $v$ to be $v_\text{new}$. This process iterates through all new vertices and their neighbors.

\vspace{-5pt}
\subsection{\sirrt}
\label{sec:sirrt}

Based on the subroutines described above, \sirrt performs as described in Alg.~\ref{alg:sirrt}. 
% initialization
An SI map $\mathcal{M}$ is initialized to be empty and then updated to add the SIs at $\vt{q}_\text{start}$ and $\vt{q}_\text{goal}$ (lines~\ref{alg3line:init1}--\ref{alg3line:init3}). \textsc{GetSafeIntervals} searches $\mathcal{C}$ over time to find all SIs during which the given configuration is collision-free. 
% lower bound of the goal node
The robot must reach $\vt{q}_\text{goal}$ after all dynamic obstacles have cleared the area so as to the robot can remain at the goal indefinitely. Thus, the lower bound of the earliest arrival time at the goal node must exceed $t_{\text{lb}}$, the start of the last SI at $\vt{q}_\text{goal}$ (line~\ref{alg3line:init4}).
% init start and goal vertex
Initialization of the vertices of the start and goal configurations follows (lines~\ref{alg3line:init_start}--\ref{alg3line:init_goal}) where $v_\text{goal}.t_\text{low} = \infty$ as the arrival time at the goal has not been computed. 

% main loop
The main loop iterates for $iteration$ times (lines~\ref{alg3line:while_begin}--\ref{alg3line:while_end}). If $\vt{q}_\text{new}$ is found from \textsc{Sampling} and \textsc{Steer}, the SIs at $\vt{q}_\text{new}$ are added to $\mathcal{M}$ (line~\ref{alg3line:update}). If new vertices $\mathcal{V}_\text{new}$ for $\vt{q}_\text{new}$ are found by \textsc{ChooseParent}, \textsc{Rewire} optimizes $G$ (line~\ref{alg3line:chooseparent}--\ref{alg3line:rewire}). 
% goal condition
If (i) the configuration of $v_\text{new}$ is equal to the goal configuration (line~\ref{alg3line:goalcheck}) and (ii) the robot could arrive at the goal at a time $t$ such that $t_{\text{lb}} < t < v_\text{goal}.t_\text{low}$ (line~\ref{alg3line:intvcheck}), $v_\text{goal}$ is updated by $v_\text{new}$. 
% anytime feature
Once a goal vertex is found, Alg.~\ref{alg:sirrt} keeps optimizing $G$ using \textsc{ChooseParent} and \textsc{Rewire} until $iteration$ reaches zero. This feature enables \sirrt to balance the computational resources and solution quality.
% Return path-                             
After all iterations, if $v_\text{goal}$ has a parent, $v_\text{goal}$ is connected to $G$ so $\pi(t)$ is returned with the goal arriving time $t_\text{final} = v_\text{goal}.t_\text{low}$ (line~\ref{alg3line:checkgoalparent}--\ref{alg3line:checkgoalparentend}). Otherwise, $v_\text{goal}$ is not connected to $G$; thus, \text{NULL} is returned (line~\ref{alg3line:returnNull}).

\begin{algorithm}[tb]
{\small
\caption{\textsc{\textsc{SI-RRT$^*$}}}
\label{alg:sirrt}
\textbf{Input}: $\mathcal{C}, \mathcal{C}_\text{free}, \mathcal{O}, G, \vt{q}_{\text{start}}, \vt{q}_{\text{goal}}, d_{\text{max}}, \lambda, iteration$\\
\textbf{Output}: Trajectory $\pi(t)$ or NULL
\begin{algorithmic}[1] %[1] enables line numbers
\STATE $\mathcal{M} \leftarrow $ an empty map\label{alg3line:init1}
\STATE $\mathcal{M}[\vt{q}_\text{start}] \leftarrow$ \textsc{GetSafeIntervals}($\vt{q}_\text{start}, \mathcal{C}_\text{free}$)\label{alg3line:init2}
\STATE $\mathcal{M}[\vt{q}_\text{goal}] \leftarrow$ \textsc{GetSafeIntervals}($\vt{q}_\text{goal}, \mathcal{C}_\text{free}$) \label{alg3line:init3}
\STATE $t_{\text{lb}} \leftarrow \max\{low : (low, high) \in \mathcal{M}[\vt{q}_\text{goal}]\}$ \label{alg3line:init4}
\STATE $v_{\text{start}}.\vt{q} \leftarrow \vt{q}_{\text{start}}, v_{\text{start}}.t_{\text{low}} \leftarrow 0, v_{\text{start}}.parent \leftarrow \text{NULL}$\label{alg3line:init_start}
\STATE $v_{\text{goal}}.\vt{q} \leftarrow \vt{q}_{\text{goal}}, v_{\text{goal}}.t_{\text{low}} \leftarrow \infty, v_{\text{goal}}.parent \leftarrow \text{NULL}$\label{alg3line:init_goal}
\WHILE{$iteration > 0$} \label{alg3line:while_begin}
    \STATE $\vt{q}_{\text{rand}} \leftarrow$ \textsc{Sampling}($\lambda$, $\vt{q}_{\text{goal}}$, $\mathcal{C}$)\label{alg3line:sampling}
    \STATE $\vt{q}_{\text{new}} \leftarrow$ \textsc{Steer}($\vt{q}_{\text{rand}}$, $\mathcal{V}$)\label{alg3line:steer}
    \IF{$\vt{q}_{\text{new}} = \text{NULL}$}
        \STATE \textbf{continue}
    \ENDIF
    \STATE $\mathcal{M}[\vt{q}_\text{new}] \leftarrow $ \textsc{GetSafeIntervals}($\vt{q}_\text{new}, \mathcal{C}_\text{free}$)\label{alg3line:update}
    \STATE $\mathcal{V}_{\text{neighbor}} \leftarrow$ \textsc{GetNeighbor}($d_{\text{max}}$, $\mathcal{V}$)
    \STATE $\mathcal{V}_{\text{new}} \leftarrow$ \textsc{ChooseParent}($q_{\text{new}}$, $\mathcal{V}_{\text{neighbor}}$, $G$, $\mathcal{M}$, $\mathcal{O}$) \label{alg3line:chooseparent}
    \IF{$\mathcal{V}_{\text{new}} = \emptyset$}
        \STATE \textbf{continue}
    \ENDIF
    \STATE \textsc{Rewire}($\mathcal{V}_{\text{new}}$, $\mathcal{V}_{\text{neighbor}}$, $G$, $\mathcal{M}$, $\mathcal{O}$)\label{alg3line:rewire}
    \FOR{$v_{\text{new}}$ in $\mathcal{V}_{\text{new}}$}
        \IF{$v_{\text{new}}.\vt{q} = v_{\text{goal}}.\vt{q}$\label{alg3line:goalcheck}} 
           \IF{$t_{\text{lb}} < v_{\text{new}}.t_{\text{low}} < v_{\text{goal}}.t_{\text{low}}$\label{alg3line:intvcheck}}
                \STATE $v_{\text{goal}} \leftarrow v_{\text{new}}$
           \ENDIF
        \ENDIF
    \ENDFOR
    \STATE $iteration \leftarrow iteration - 1$
\ENDWHILE \label{alg3line:while_end}
\IF {$v_{\text{goal}}.parent \neq \text{NULL}$} \label{alg3line:checkgoalparent}
    \STATE $t_\text{final} = v_\text{goal}.t_{\text{low}}$%\textsc{ExtractPath}($v_{\text{goal}}$
    \RETURN $\pi(t)$ for $t \in [0, t_\text{final}]$
\ENDIF  \label{alg3line:checkgoalparentend}
\RETURN \text{NULL} \label{alg3line:returnNull}
\end{algorithmic}
}
\end{algorithm}

\vspace{-5pt}
\subsection{Analysis of \sirrt}
We prove the probabilistic completeness and the asymptotic optimality of \sirrt.

\begin{theorem}
    \sirrt is probabilistically complete.
\end{theorem}
\vspace{-10pt}
\begin{proof}
    Let us consider $\vt{q}_\text{new}$ and its $\mathcal{V}_\text{neighbor}$ in the roadmap $G$. For each $\iota_\text{new}$ of $\vt{q}_\text{new}$, \sirrt attempts to create $v_\text{new}$ and connect it to each $v \in \mathcal{V}_\text{neighbor}$. A connection between $v_\text{new}$ and $v$ is established unless either of the following conditions is met:\vspace{-5pt}
    \begin{itemize}
        \item (Condition 1) Non-overlapping safe intervals: If the interval resulting from linearly translating the SI of $v$ by the moving time from $v.\vt{q}$ to $\vt{q}_\text{new}$ does not overlap with $\iota_\text{new}$, the robot will inevitably collide with an obstacle, regardless of its departure time from $v$.
        \item (Condition 2) Unavoidable collisions: Even with overlapping safe intervals, if all trajectories from $v$ to $\vt{q}_\text{new}$ within $\iota_\text{new}$ collide with dynamic obstacles regardless of departure time, no feasible trajectory exists.
    \end{itemize}\vspace{-4pt}

    Based on these conditions, $v_\text{new}$ is added to the tree if there exists at least one collision-free trajectory from any neighbor $v$ to $v_\text{new}$ within $\iota_\text{new}$, satisfying both the time overlap and dynamic obstacle avoidance conditions.
    For each sampled configuration $\vt{q}_\text{new}$, \sirrt attempts to create $\mathcal{V}_{\text{new}}$ for all of its SIs, ensuring comprehensive exploration of all feasible time windows. As sampling continues indefinitely, the tree progressively covers all reachable configurations and their associated SIs. This expansion results in a representation of all combinations of collision-free spatial and temporal configurations. The continuous expansion of $G$ eventually ensures that \sirrt will connect $v_\text{start}$ to $v_\text{goal}$ if a feasible path exists.
\end{proof}

\begin{theorem}
    \sirrt is asymptotically optimal.
\end{theorem}
\vspace{-10pt}
\begin{proof}
    First, \sirrt maintains the earliest arrival time at each vertex in $G$. Therefore, the earliest arrival time at $v_\text{goal}$, which is $v_\text{goal}.t_\text{low}$ (i.e., $t_\text{final}$), represents the minimum time to reach the goal within the structure of the current $G$. We can prove the asymptotic optimality of \sirrt based on the following two observations:\vspace{-5pt}
    \begin{enumerate} 
        \item Representation optimality\footnote{The notion of being \textit{representation-optimal}, also outlined in~\cite{cbs-mp}, entails achieving an optimal solution within a given representation of the solution space. The solution space is not necessarily identical to the representation.}: Once a solution is found, the trajectory $\pi(t)$ for $t \in [0, t_\text{final}]$ is optimal within the current representation of $G$.
        \item Convergence to configuration space: As the number of samples increases, the set of vertices $\mathcal{V}$ in roadmap $G$ increasingly approximates $\mathcal{C}_\text{free}$.
    \end{enumerate}

    Under infinite uniform sampling, $G$ asymptotically covers all configurations in $\mathcal{C}_\text{free}$. Consequently, the representation converges to the solution space. Given that \sirrt is representation-optimal, the solution found from $G$ approaches optimal as the number of samples approaches infinity.
\end{proof}

\vspace{-10pt}
\subsection{An extension for kinodynamic constraints of robots}
In real-world scenarios, physical robots are subject to various kinodynamic constraints. Our method can address these constraints by incorporating a local planner that finds trajectories under kinodynamic constraints between two states. 
% Any local planner is acceptable if the planner ensures that the local collision-free path meets the earliest arrival time at each destination state of a robot. 
Any local planner is acceptable if the planner can generate a collision-free trajectory connecting two states and guarantee the earliest possible arrival time at each destination state for a robot.

As an example, we implement \textit{bang-bang transform}~\cite{bangbang} for the local planner. %This transform is a variant of \textit{bang-bang control}, adapted for scenarios where zero velocity is assumed at all nodes. 
With the planner, the robot accelerates at the maximum rate for a certain duration, navigates with its maximum velocity, and then decelerates until it reaches the next state with zero velocity. For each pair of states, we set the control input as $a_\text{max} = -a_\text{min} = 1$, calculate the velocity vector $\vt{v} = \vt{q'} - \vt{q}$, normalize it as $\hat{\vt{v}} = \vt{v} / |\vt{v}|$, and compute $s = \max_k(|\hat{v_k}|)$, $a_k = \hat{v_k}/s$, and $t = \sqrt{s|\vt{v}|}$, where $k$ represents each dimension of the task space. It is worth noting that this implementation is merely an example of local planners that can handle kinodynamic constraints as various methods are available.

Furthermore, our method can deal with robots of different sizes and shapes. To show the capability, we randomly set different sizes for the robots in our implementation.\footnote{Robots with noncircular shapes can be represented by the circles circumscribing them.} The details of the implementation for kinodynamics and nonregular sizes can be seen in the video clip and the source code provided as supplementary materials.

\vspace{-5pt}
\section{Conflict resolution for multiple robots}
\label{sec:multi}
%Although \sirrt enables each robot to find its own collision-free trajectory, conflicts may occur between the trajectories of multiple robots. 

As the high level manages conflicts between individual collision-free trajectories generated by the low-level \sirrt, we integrate PP and CBS with \sirrt, called \sipp and \sicbs, respectively. While \sipp is more efficient, \sicbs produces solutions with better quality, so one can choose either of them according to the requirements of the task.

\vspace{-5pt}
\subsection{Safe Interval Continuous-space PP (SI-CPP)}
PP is a general approach based on a given, fixed priority of robots. Each robot avoids collisions with the trajectories of the higher-priority robots. Lower-priority robots treat the trajectories of higher-priority robots as dynamic obstacles when they plan the trajectories. Although PP is simple to implement and efficient, it is not complete. \sipp also inherits this property.

\vspace{-5pt}
\subsection{Safe Interval Continuous-space CBS (SI-CCBS)}
CBS is an optimal and complete two-level algorithm. The high-level of \sicbs, following the structure of CBS, performs a best-first search in the Constraint Tree (CT). We adopt the cost function of the greedy variant of CBS proposed in~\cite{ecbs}, which is the number of conflicts between trajectories for fast search. To handle continuous space, we modify the definitions of conflicts and constraints.

\sicbs begins by inserting a root node with an empty constraint set into the search frontier $OPEN$. Each CT node contains the cost, individual trajectories, and constraints. The search iterates by expanding the node with the minimum cost from $OPEN$. The expanded node is eliminated from $OPEN$. If the expanded node has no conflict, the search terminates with a conflict-free solution. If $OPEN$ is empty, the search fails to find a solution.

When expanding a CT node, conflicts among all trajectories are found and used to determine the cost of the node. To create the constraint set for a child CT node, we add a new constraint to the constraint set of the parent node. This new constraint is derived from a conflict  (we choose the conflict with the smallest $t$) between the trajectories $\pi^i(t)$ and $\pi^j(t)$ of robots $i$ and $j$, respectively. Subsequently, \sirrt finds new trajectories that abide by the updated constraint set. A child node updates its cost and trajectory set accordingly and is inserted into $OPEN$.

\vspace{-5pt}
\section{Experiments}
\label{sec:experiments}
 First, we evaluate the performance of \sirrt for single-robot instances. Then, a comprehensive comparison of multi-robot planners follows to compare \sipp and \sicbs with others that are capable of handling continuous space and time. We have four test environments with different obstacle profiles: \textsf{\small Circ10}, \textsf{\small Circ20}, \textsf{\small Rect10}, and \textsf{\small Rect20} as shown in Fig.~\ref{fig:env}. The size of all environments is $40\,m \times 40\,m$ where the radius of all robots is $0.5\,m$. \textsf{\small Circ10} and \textsf{\small Circ20} are filled with circular obstacles with different densities of the obstacles (10\% and 20\% of the entire space, respectively). Likewise, \textsf{\small Rect10} and \textsf{\small Rect20} are with rectangular obstacles. For each environment, we generate a set of $50$ unique instances while randomizing the start and goal locations as well as the size and location of the obstacles. The parameter settings are as follows: $\lambda = 0.1, d_{\text{max}} = 5\,m$, $iteration = 1500$, and the maximum velocity of the robots is $0.5\,m/s$. The system has an AMD Ryzen 5800X 3.8GHz CPU and 32G RAM, and all algorithms except for GT are implemented in C++17. GT is implemented using Python 3.8 because of its dependency on PyTorch.

\vspace{-5pt}
\subsection{Single-robot path planning}
\label{sec:single-ex}
% the test environment
We compare \sirrt to \strrt, one of the state-of-the-art sampling-based planners in continuous space with dynamic obstacles. \strrt has been shown to outperform algorithms like RRT-Connect and RRT$^*$ in dynamic environments~\cite{strrtstar}. For comparison, we set up an environment with $60$ dynamic obstacles with the same size and shape as the robot. The trajectories of the obstacles are randomly generated but known to both algorithms. We use the earliest arrival time at $\vt{q}_\text{goal}$ as our primary performance metric, reflecting search tree quality and trajectory efficiency. Specifically, a lower value indicates a better solution. To evaluate sample efficiency, we measure this metric while increasing the number of samples.

\begin{figure}[t]
\captionsetup{skip=0pt}
\centering
\includegraphics[width=0.88\linewidth]{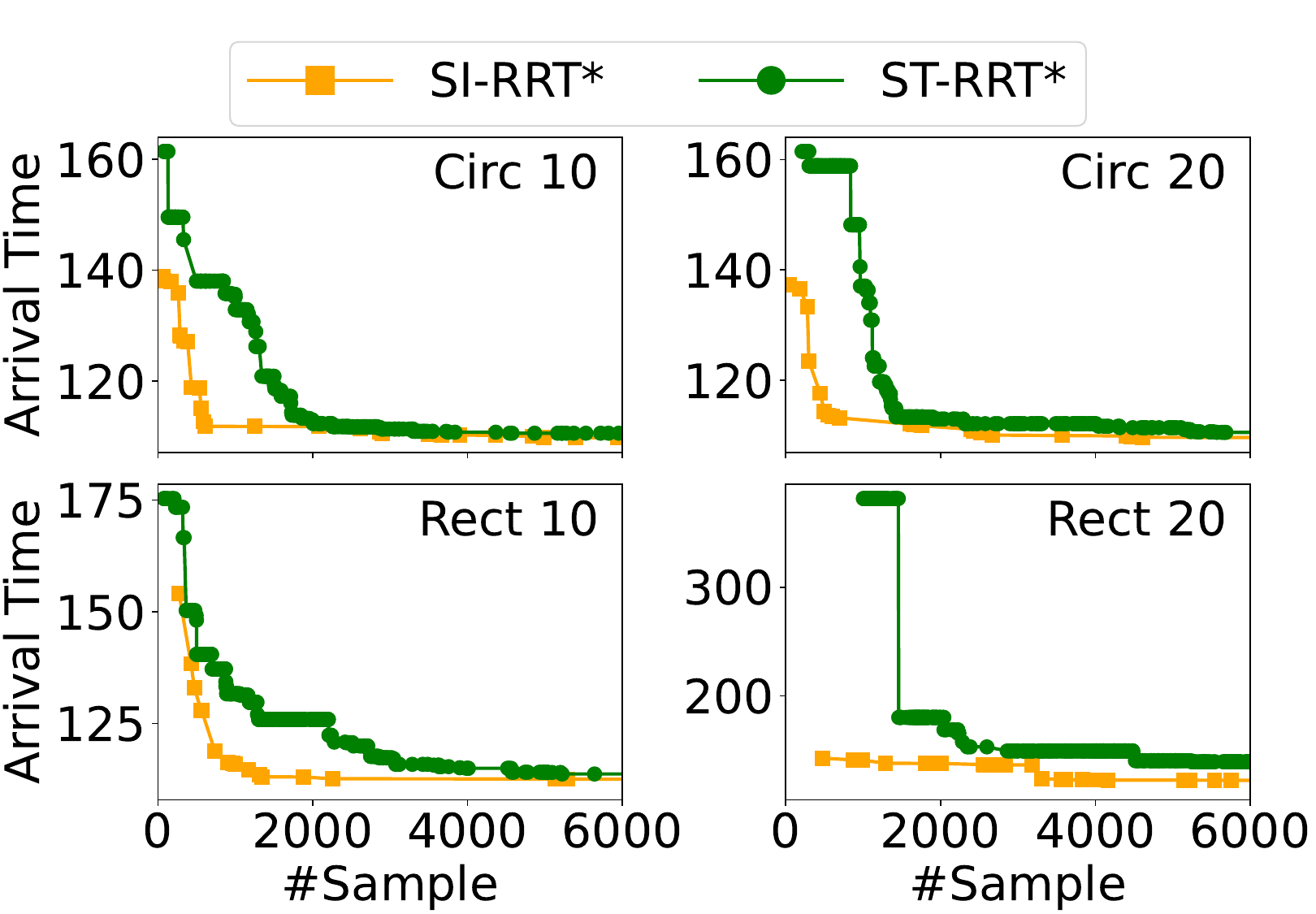}%
\caption{A comparison between \sirrt (yellow) and \strrt (green) for single-robot path planning in the four test environments. The earliest arrival time at the goal of \sirrt decreases faster than \strrt, indicating that \sirrt can find better trajectories faster.}\vspace{-12pt}
\label{fig:single}
\end{figure}

% test result and test analysis
As shown in Fig.~\ref{fig:single}, \sirrt decreases the earliest arrival time at $\vt{q}_\text{goal}$ than \strrt consistently across all environments. This suggests \sirrt discovers high-quality trajectories more quickly. 
%Although the performance gap narrows with increasing samples, \sirrt maintains faster convergence to lower arrival times throughout.
The faster convergence of \sirrt stems from the dimensionality of the sampling space. \sirrt samples only in the spatial dimension while \strrt samples from both spatial and temporal dimensions. Thus, \sirrt can cover more spatial configurations than \strrt with the same number of samples. 
The performance gap increases in environments with more obstacles. 
%If \strrt rejects a sample owing to the obstacles in the spatial dimension, the rejected sample may expand the temporal dimension without growing $G$. Thus, the resulting sampling space becomes larger without any progress in searching for the solution. % New커멘트 : 기존에 썼던 내용이 모호하게 적혀있어서 Ln 601로 새로 썼으니 확인 바랍니다. 
% rejected sample이 temporal dimension을 확장한다는게 어색해서 아래와 같이 다시 작성해봤습니다. + sirrt의 특성에 대해서도 설명이 필요할 것 같아서 그 부분도 추가작성했습니다. 
% 내일 discussion 할 내용
% - it tends to expand more in the temporal dimension -> "more"에 대한 근거가 없음 
% - inefficient growth of the search space -> growth가 inefficient하다는 표현이 이상
If \strrt rejects a sample owing to obstacles in the spatial dimension, resampling expands both the spatial and temporal dimensions of the search space, even though valid trajectories might exist without expanding the temporal dimension. The increased search space incurs inefficiency as more samples are needed to explore the spacial dimension. In contrast, \sirrt does not sample from the temporal dimension so can keep the search space smaller than \strrt.

\vspace{-5pt}
\subsection{Multi-robot path planning}
\label{sec:multi-ex}
% metric introduction
The metrics are the flowtime $\kappa$, makespan $\tau$, and success rate. In congested environments, the importance of flowtime and makespan becomes evident as robots often need to wait to avoid conflicts with one another. The success rate is defined as the number of instances where a solution is found within five minutes out of all 50 test instances.

% Compare Algorithms
We compare \sipp and \sicbs to \strrt with \strrtpp, SSSP, and GT across the four test environments. We vary the number of robots up to 160.
% the different features of the algorithms
SSSP and GT operate in discrete time steps, requiring synchronization because robots must wait at waypoints until other robots arrive. In contrast, \strrtpp and our methods operate in continuous time, so synchronization is unnecessary.

\begin{figure}[t]
\captionsetup{skip=0pt}
\centering
\includegraphics[width=0.99\linewidth]{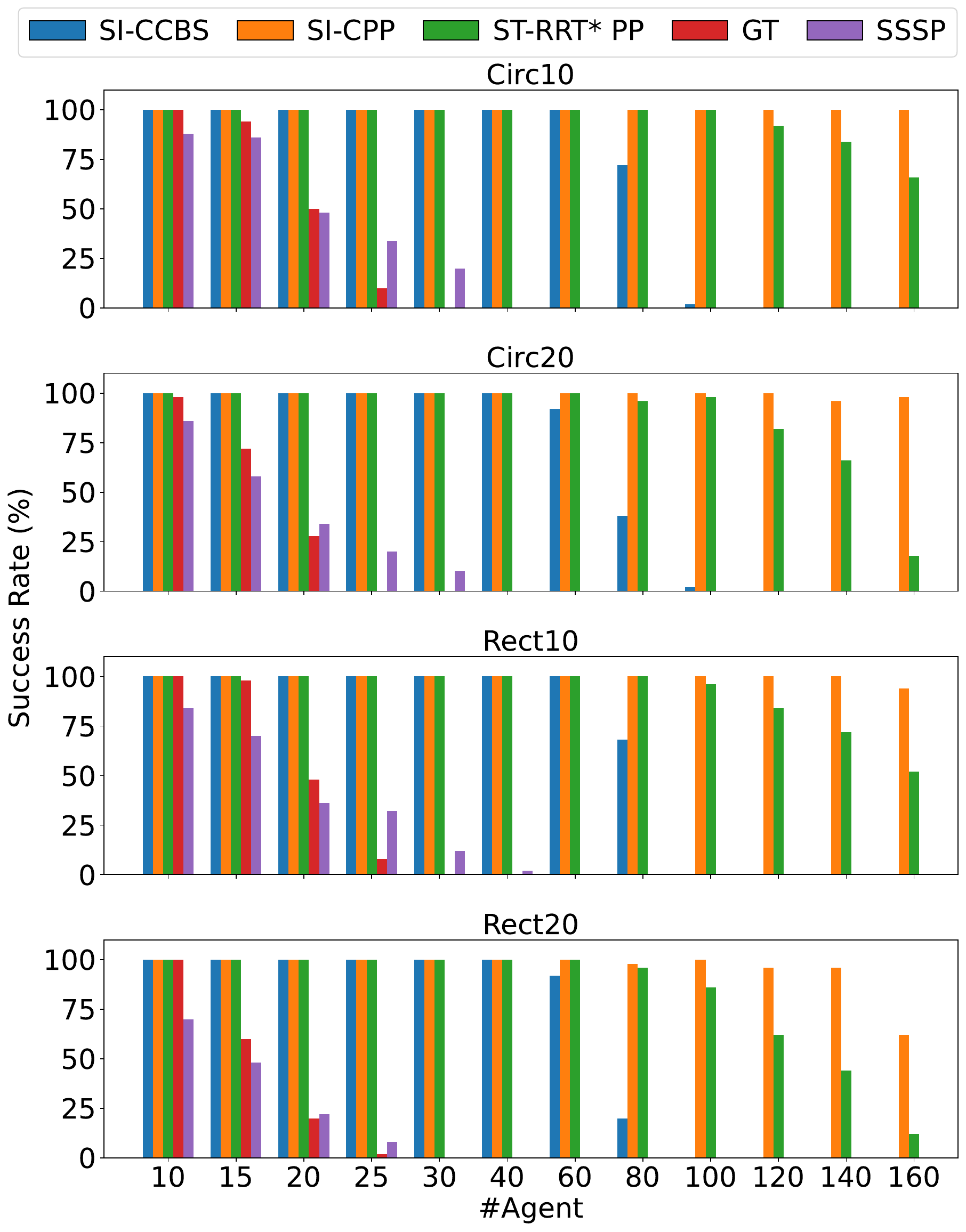}\
\caption{The success rates of compared methods under five minutes of the time limit. The proposed method \sipp succeeds in almost all instances, even in congested environments. As the advantage of \sicbs is in its solution quality, its success rates are lower than \sipp and \strrtpp but significantly higher than SSSP and GT.}\vspace{-12pt}
\label{fig:success}
\end{figure}

% success rate
The success rates shown in Fig.~\ref{fig:success} demonstrate that \sipp, \sicbs, and \strrtpp are scalable. They maintain nearly 100\% success rates for up to 60 robots. \sipp is notably scalable as it achieves 96\% of the success rate even with 140 robots in the most challenging environment, \textsf{\small Rect20}. \sicbs maintains high success rates up to 60 robots but shows a decline for 80 robots in denser environments. \strrtpp is efficient, but its success rate declines with more than 100 robots. On the other hand, the scalability of SSSP and GT is limited. Their success rates drop significantly with more than 20 robots across all environments, especially in \textsf{\small Circ20} and \textsf{\small Rect20}. Both methods fail to find solutions for problem instances beyond 40 robots.

\begin{figure*}[t!]
\captionsetup{skip=0pt}
\centering
\includegraphics[width=0.99\linewidth]{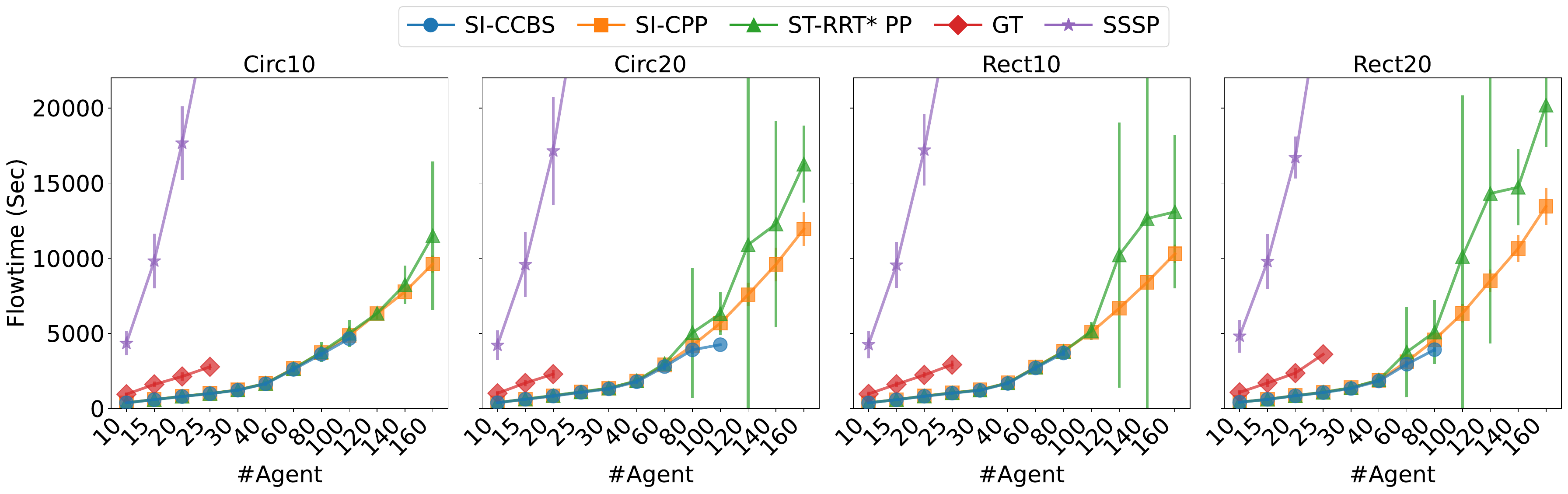}
\caption{Compared flowtime where the standard deviation is displayed by the error bar in \textsf{\small Circ10}, \textsf{\small Circ20}, \textsf{\small Rect10} and \textsf{\small Rect20}}\vspace{-12pt}
\label{fig:flowtime}
\end{figure*}

\begin{figure}[t]
\captionsetup{skip=0pt}
\centering
\includegraphics[width=0.99\linewidth]{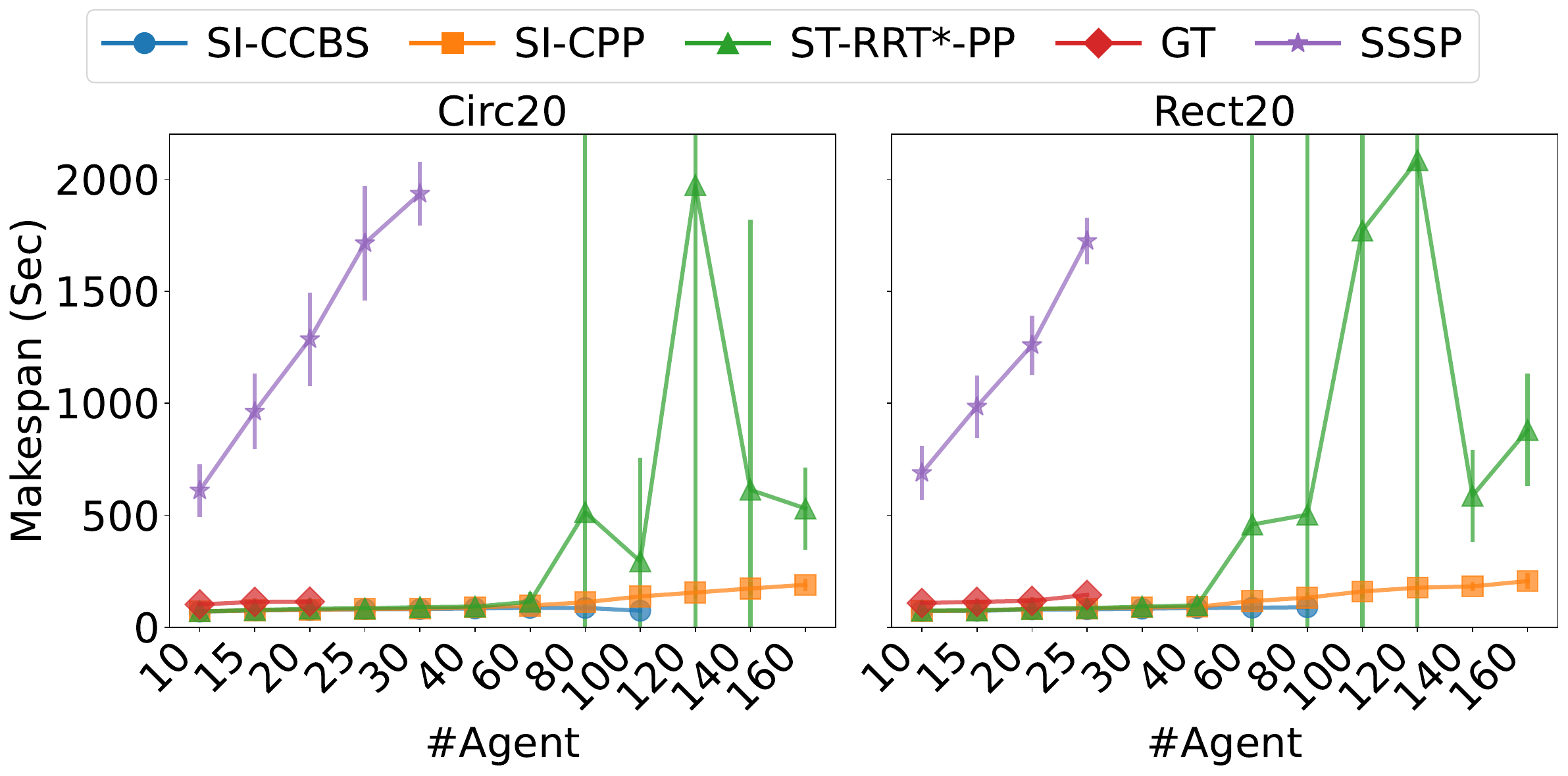}%
\caption{Compared makespans in \textsf{\small Circ20} and \textsf{\small Rect20}}\vspace{-12pt}
\label{fig:makespan}
\end{figure}

% solution quality
Figs.~\ref{fig:flowtime} and \ref{fig:makespan} show comparisons of the solution quality of the compared algorithms for different robot team sizes.
% SSSP and GT
First, SSSP and GT exhibit significant performance drops as the problem size increases. The sequential movement of SSSP (discussed in the related work section) results in a rapid increase in flowtime. While GT shows a better performance than SSSP, its learning-based heuristic appears to be less effective in complex environments. Both methods yield substantially high flowtime and makespan values in most cases.
% Compared to ST-RRT* and SI-RRT* 
From these results of SSSP and GT, we decide to focus on comparing our proposed methods (\sipp and \sicbs) with \strrtpp in the subsequent experiments.

% 60 robots
For 60 robots, where all three methods maintain high success rates, \sicbs shows notable improvements. It achieves reductions of 6.48\% and 21.4\% in flowtime compared to \strrtpp in \textsf{\small Circ20} and \textsf{\small Rect20} environments, respectively.
% 100 robots
As we increase the number of robots to 100, the performance gap increases more significantly. \sipp reduces the flowtime 9.8\% and 37.3\% compared to \strrtpp in \textsf{\small Circ20} and \textsf{\small Rect20} environments, respectively.

% 160 robots
Since we measure the flowtime and makespan for those only successful instances, they could be biased when the success rate is low, as only easy instances are likely to succeed. In instances with 160 robots, \strrtpp experiences a significant decline in the success rate. Nevertheless, \sipp outperforms \strrtpp in terms of the flowtime and makespan. Particularly, \sipp shows significant reductions in the flowtime in \textsf{\small Circ20} and  \textsf{\small Rect20}, which are 26.5\% and 33.3\%, respectively.

% Makespan
Furthermore, our methods show significant improvements in terms of makespan, particularly in large-size instances. With 60 robots, \sicbs reduces the makespan up to 23.9\% in \textsf{\small Circ20} and 80.9\% in \textsf{\small Rect20} compared to \strrtpp. For 120 robots, \sipp shows more significant reductions up to 92.12\% in \textsf{\small Circ20} and 91.51\% in \textsf{\small Rect20}.

% std
The standard deviations of the measured metrics provide additional insights. As the number of robots increases, \strrtpp exhibits a substantial growth in the standard deviation while \sipp and \sicbs maintain relatively low variability. These observations suggest that our methods are more consistent and reliable across diverse scenarios.

% why is our algorithm better?
The superior performance of \sirrt in the solution quality can arise from its novel approach to edge connections. Unlike \strrt, which samples in both spatial and temporal C-spaces, \sirrt samples only in the spatial C-space. In \strrt, connecting vertex sampled in both space and time implicitly imposes velocity constraints due to fixed spatial and temporal distances between nodes. In contrast, the spatial-only sampling of \sirrt allows for determining unconstrained velocities between states, enabling the calculation of the earliest feasible arrival time at each destination. Consequently, \sirrt can generate more time-efficient trajectories, potentially minimizing overall travel time across the robot team.

% summary
In summary, \sicbs offers the highest quality solutions regarding flowtime and makespan, with reasonable scalability of up to 60 robots. \sipp balances the solution quality and scalability, maintaining high performance even for 160 robots in challenging environments. While \strrtpp shows comparable success rates in many instances, it is outperformed by our methods, particularly for larger team sizes and denser environments. Since we are unable to report all numbers from the experiments owing to the space limit, we attach them in the supplementary material.

\vspace{-5pt}
\section{Conclusion}
We proposed a two-level approach for addressing the multi-robot path planning problem in continuous space to generate collision-free trajectories efficiently. Based on the sample-efficient \sirrt, two variants \sipp and \sicbs for MRPP were presented. Experiments demonstrated that \sipp outperforms a state-of-the-art method in terms of the success rate, which shows the scalability of \sipp given a time budget for computation. \sipp also shows a significantly smaller makespan. \sicbs outperforms the state-of-the-art method regarding the solution quality while not sacrificing the scalability significantly. As a future direction, we plan to run experiments with physical robots. 

% \bibliography{aaai25}
\bibliography{references}

\end{document}

\title{\LARGE{Supplementary Material}}

\maketitle

\IEEEpeerreviewmaketitle
In this document, we provide (i) detailed explanations of the proposed algorithms that were omitted from the manuscript (\textsf{ChooseParent} and \textsf{Rewire}) due to space constraints, (ii) numerical results of the experiments, and (iii) additional experimental validations conducted in dynamic simulations. The accompanying video clip includes test instances performed in both the test environment described in the manuscript and the dynamic simulation environment.

\section{Detailed explanations of \textsf{ChooseParent} and \textsf{Rewire}}
In \textsc{ChooseParent} (Alg.~\ref{alg:chooseparent}), a new set of vertices, $\mathcal{V}_\text{new}$, is initialized as empty (line~\ref{alg1line:initialvset}).
Alg.~\ref{alg:chooseparent} iterates over all safe intervals associated with $\vt{q}_\text{new}$, attempting to create a new vertex for each interval (line~\ref{alg1line:fori}). For each safe interval $\iota_\text{new}$, a new vertex $v_\text{new}$ is initialized with its configuration set to $\vt{q}_\text{new}$, its earliest arrival time $t_\text{low}$ set to infinity, and its parent vertex set to \text{NULL} (line~\ref{alg1line:initialvnew}). The algorithm then iterates through $\mathcal{V}_\text{neighbor}$ (line~\ref{alg1line:forv}) to determine if any of them can serve as the parent vertex for $v_\text{new}$. For each neighboring vertex $v$, the algorithm checks if there exists a collision-free trajectory from $v$ to $v_\text{new}$ within the safe interval $\iota_\text{new}$. If such a collision-free trajectory exists, the earliest possible arrival time at $v_\text{new}$ is calculated and stored in the variable $low$ (line~\ref{alg1line:low}).

\begin{algorithm}
{\small
    \caption{\textsc{ChooseParent}}
    \label{alg:chooseparent}
    \textbf{Input}: $\vt{q}_{\text{new}}$, $\mathcal{V}_{\text{neighbor}}$, $G$, $\mathcal{M}$, $\mathcal{O}$\\
    \textbf{Output}: $\mathcal{V}_{\text{new}}$
    \begin{algorithmic}[1] %[1] enables line numbers
        \STATE $\mathcal{V}_{\text{new}} \leftarrow \emptyset$~\label{alg1line:initialvset}
        \FOR{$\iota_\text{new}$ in $\mathcal{M}(\vt{q}_{\text{new}})$~\label{alg1line:fori}}
            \STATE $v_{\text{new}}.\vt{q} \leftarrow \vt{q}_{\text{new}}$, $v_{\text{new}}.t_\text{low} \leftarrow \infty$, $v_{\text{new}}.parent = \text{NULL}$~\label{alg1line:initialvnew}
            \FOR{$v$ in $\mathcal{V}_{\text{neighbor}}$~\label{alg1line:forv}}
                \STATE $low \leftarrow$ the earliest arrival time of $v_{\text{new}}.\vt{q}$ within $[\iota_\text{new}.low, \iota_\text{new}.high)$ without colliding with $\mathcal{O}$ when the robot moves from $v$, considering the execution time required to move from $v$ to $v_{\text{new}}$  \label{alg1line:low}
                \IF{$low$ does not exist}
                    \STATE \textbf{continue}\label{alg1line:continue}
                \ENDIF
                \IF{$low < v_{\text{new}}.t_{\text{low}}$} \label{alg1line:update1}
                    \STATE $v_{\text{new}}.t_{\text{low}} \leftarrow low$
                    \STATE $v_{\text{new}}.parent \leftarrow v$
                \ENDIF\label{alg1line:update2}
            \ENDFOR~\label{alg1line:endforv}
            \IF{$v_{\text{new}}.parent \neq \text{NULL}$} \label{alg1line:update3}
                \STATE $\mathcal{V}_{\text{new}} \leftarrow \mathcal{V}_{\text{new}} \cup v_\text{new}$
                \STATE $\mathcal{V} \leftarrow \mathcal{V} \cup v_\text{new}$
                \STATE $\mathcal{E} \leftarrow \mathcal{E} \cup (v_\text{new}, v_\text{new}.parent)$
            \ENDIF~\label{alg1line:update4}
        \ENDFOR~\label{alg1line:endforI}
        \STATE \textbf{return} $\mathcal{V}_{\text{new}}$ 
    \end{algorithmic}
}
\end{algorithm}

\begin{algorithm}
{\small
    \caption{\textsc{Rewire}}
    \label{alg:rewire}
    \textbf{Input}: $\mathcal{V}_{\text{new}}$, $\mathcal{V}_{\text{neighbor}}$, $G$, $\mathcal{M}$, $\mathcal{O}$\\
    \textbf{Output}: \text{None}
    \begin{algorithmic}[1] %[1] enables line numbers
        \FOR{$v_{\text{new}}$ in $\mathcal{V}_{\text{new}}$} \label{alg2line:forvnew}
            \FOR{$v$ in $\mathcal{V}_{\text{neighbor}}$} \label{alg2line:forv}
                \FOR{$\iota_\text{neighbor}$ in $\mathcal{M}(v.\vt{q})$} \label{alg2line:fori}
                    \STATE $low \leftarrow$ the earliest arrival time of $v.\vt{q}$ within $[\iota_{\text{neighbor}}.low, \iota_{\text{neighbor}}.high)$ without colliding with $\mathcal{O}$ when the robot moves from $v_{\text{new}}$, considering the execution time required to move from $v$ to $v_{\text{new}}$\label{alg2line:low}
                    \IF{$low$ does not exist}
                        \STATE \textbf{continue}
                    \ENDIF
                    \IF{$low < v.t_\text{low}$\label{alg2line:update}}
                        \STATE $v.t_{\text{low}} \leftarrow low$ \label{alg2line:if_update1}
                        \STATE $\mathcal{E} \leftarrow \mathcal{E} \setminus (v, v.parent)$ \label{alg2line:if_update1}
                        \STATE $v.parent \leftarrow v_{\text{new}}$ \label{alg2line:if_update3}
                        \STATE $\mathcal{E} \leftarrow \mathcal{E} \cup (v, v_\text{new})$
                        \label{alg2line:if_update2}
                    \ENDIF\label{alg2line:updateend}
                \ENDFOR
            \ENDFOR
        \ENDFOR
    \end{algorithmic}
}
\end{algorithm}

% Detailed explanation of how to calculate low
In the left of Fig.~3 of the manuscript, the robot can stay at configuration $v.\vt{q}$ from time $v.t_{\text{low}}$ until time $\iota.high$ without collisions. The safe interval of $v_\text{new}.\vt{q}$ (green) is $[\iota_\text{new}.low, \iota_\text{new}.high)$. In the right of Fig.~3 in the manuscript, the bold arrow represents a potential trajectory for the robot to move from $v.\vt{q}$ to $v_\text{new}.\vt{q}$ as early as possible while avoiding collisions with the dynamic obstacle (blue). 
Thus, the calculated earliest arrival time at configuration ($v_\text{new}.\vt{q}$) along this collision-free trajectory is assigned to the variable $low$.

If the safe interval $\iota_\text{new}$ of $v_\text{new}.\vt{q}$ does not overlap with the time range in which the robot can reach $v_\text{new}.\vt{q}$ from $v$, no feasible trajectory exists between them. Even with overlapping time ranges, a feasible trajectory may not exist if the trajectory from $v$ to $v_\text{new}.\vt{q}$ inevitably collides with a dynamic obstacle, regardless of departure time. In both cases, a valid earliest arrival time $low$ cannot be determined. When $low$ is not found, the algorithm moves to the next neighboring vertex in $\mathcal{V}_\text{neighbor}$ (line~\ref{alg1line:continue}).
% In case there is low
If a valid $low$ is found and it is lower than the current earliest arrival time for $v_\text{new}$, both the arrival time and parent vertex of $v_\text{new}$ are updated (lines~\ref{alg1line:update1}--\ref{alg1line:update2}). Iterating over $\mathcal{V}_\text{neighbor}$ ensures $v_\text{new}$ is assigned the earliest possible arrival time within the current tree structure. If $v_\text{new}$ has a parent, indicating a viable path, it is added to both $\mathcal{V}_\text{new}$ and $G$ (lines~\ref{alg1line:update3}--\ref{alg1line:update4}).

\textsc{Rewire} (Alg.~\ref{alg:rewire}) optimizes $G$ by reconnecting edges when better trajectories are possible, using the set $\mathcal{V}_\text{new}$ from \textsc{ChooseParent}. For each $v_\text{new} \in \mathcal{V}_\text{new}$, it evaluates whether making $v_\text{new}$ the parent of any $v \in \mathcal{V}_\text{neighbor}$ would lower earliest arrival time of $v$. This is done by computing the potential earliest arrival time at $v$ when reaching it via $v_\text{new}$ (line~\ref{alg2line:low}). If a valid earliest arrival time $low$ is found and earlier than the current earliest arrival time $v.t_\text{low}$, then the algorithm updates $v.t_\text{low}$ to $low$ and reassigns the parent of $v$ to be $v_\text{new}$ (lines~\ref{alg2line:update}--\ref{alg2line:updateend}). This process iterates through all new vertices and their neighbors.

\section{Numerical results of the experiments}

In the manuscript, we were not able to include exact measured values due to the space constraint. In Table~\ref{tab:exp}, all numerical results are provided for all environments \textsf{Circ10}, \textsf{Rect10}, \textsf{Circ20}, and \textsf{Rect20}. 

\begin{table}[h!]
\centering
    \caption{Measured metrics in the experiments of the manuscript}\label{tab:exp}
\resizebox{\columnwidth}{!}{%
\begin{subtable}[t]{\textwidth}
\centering
\captionsetup{skip=1pt}
\caption{Environments \textsf{Circ10} and \textsf{Rect10}}
\begin{tabular}{|cc|ccccc|ccccc|}
\hline
\multicolumn{2}{|c|}{Environment} &
  \multicolumn{5}{c|}{\textsf{Circ10}} &
  \multicolumn{5}{c|}{\textsf{Rect10}} \\ \hline
\multicolumn{1}{|c|}{Metric} &
  \#robot &
  \multicolumn{1}{c|}{SI-CCBS} &
  \multicolumn{1}{c|}{SI-CPP} &
  \multicolumn{1}{c|}{\strrtpp} &
  \multicolumn{1}{c|}{GT} &
  SSSP &
  \multicolumn{1}{c|}{SI-CCBS} &
  \multicolumn{1}{c|}{SI-CPP} &
  \multicolumn{1}{c|}{\strrtpp} &
  \multicolumn{1}{c|}{GT} &
  SSSP \\ \hline
\multicolumn{1}{|c|}{\multirow{8}{*}{Success rate (\%)}} &
  20 &
  \multicolumn{1}{c|}{100} &
  \multicolumn{1}{c|}{100} &
  \multicolumn{1}{c|}{100} &
  \multicolumn{1}{c|}{50} &
  48 &
  \multicolumn{1}{c|}{100} &
  \multicolumn{1}{c|}{100} &
  \multicolumn{1}{c|}{100} &
  \multicolumn{1}{c|}{48} &
  36 \\ \cline{2-12} 
\multicolumn{1}{|c|}{} &
  40 &
  \multicolumn{1}{c|}{100} &
  \multicolumn{1}{c|}{100} &
  \multicolumn{1}{c|}{100} &
  \multicolumn{1}{c|}{0} &
  0 &
  \multicolumn{1}{c|}{100} &
  \multicolumn{1}{c|}{100} &
  \multicolumn{1}{c|}{100} &
  \multicolumn{1}{c|}{0} &
  2 \\ \cline{2-12} 
\multicolumn{1}{|c|}{} &
  60 &
  \multicolumn{1}{c|}{100} &
  \multicolumn{1}{c|}{100} &
  \multicolumn{1}{c|}{100} &
  \multicolumn{1}{c|}{0} &
  0 &
  \multicolumn{1}{c|}{100} &
  \multicolumn{1}{c|}{100} &
  \multicolumn{1}{c|}{100} &
  \multicolumn{1}{c|}{0} &
  0 \\ \cline{2-12} 
\multicolumn{1}{|c|}{} &
  80 &
  \multicolumn{1}{c|}{72} &
  \multicolumn{1}{c|}{100} &
  \multicolumn{1}{c|}{100} &
  \multicolumn{1}{c|}{0} &
  0 &
  \multicolumn{1}{c|}{68} &
  \multicolumn{1}{c|}{100} &
  \multicolumn{1}{c|}{100} &
  \multicolumn{1}{c|}{0} &
  0 \\ \cline{2-12} 
\multicolumn{1}{|c|}{} &
  100 &
  \multicolumn{1}{c|}{2} &
  \multicolumn{1}{c|}{100} &
  \multicolumn{1}{c|}{100} &
  \multicolumn{1}{c|}{0} &
  0 &
  \multicolumn{1}{c|}{0} &
  \multicolumn{1}{c|}{100} &
  \multicolumn{1}{c|}{96} &
  \multicolumn{1}{c|}{0} &
  0 \\ \cline{2-12} 
\multicolumn{1}{|c|}{} &
  120 &
  \multicolumn{1}{c|}{0} &
  \multicolumn{1}{c|}{100} &
  \multicolumn{1}{c|}{92} &
  \multicolumn{1}{c|}{0} &
  0 &
  \multicolumn{1}{c|}{0} &
  \multicolumn{1}{c|}{100} &
  \multicolumn{1}{c|}{84} &
  \multicolumn{1}{c|}{0} &
  0 \\ \cline{2-12} 
\multicolumn{1}{|c|}{} &
  140 &
  \multicolumn{1}{c|}{0} &
  \multicolumn{1}{c|}{100} &
  \multicolumn{1}{c|}{84} &
  \multicolumn{1}{c|}{0} &
  0 &
  \multicolumn{1}{c|}{0} &
  \multicolumn{1}{c|}{100} &
  \multicolumn{1}{c|}{72} &
  \multicolumn{1}{c|}{0} &
  0 \\ \cline{2-12} 
\multicolumn{1}{|c|}{} &
  160 &
  \multicolumn{1}{c|}{0} &
  \multicolumn{1}{c|}{100} &
  \multicolumn{1}{c|}{66} &
  \multicolumn{1}{c|}{0} &
  0 &
  \multicolumn{1}{c|}{0} &
  \multicolumn{1}{c|}{94} &
  \multicolumn{1}{c|}{52} &
  \multicolumn{1}{c|}{0} &
  0 \\ \hline
\multicolumn{1}{|c|}{\multirow{8}{*}{Flowtime (sec)}} &
  20 &
  \multicolumn{1}{c|}{806.16} &
  \multicolumn{1}{c|}{823.20} &
  \multicolumn{1}{c|}{819.38} &
  \multicolumn{1}{c|}{2139.0} &
  17662.1 &
  \multicolumn{1}{c|}{821.35} &
  \multicolumn{1}{c|}{839.85} &
  \multicolumn{1}{c|}{835.92} &
  \multicolumn{1}{c|}{2240.9} &
  17202.5 \\ \cline{2-12} 
\multicolumn{1}{|c|}{} &
  40 &
  \multicolumn{1}{c|}{1653.05} &
  \multicolumn{1}{c|}{1683.74} &
  \multicolumn{1}{c|}{1679.72} &
  \multicolumn{1}{c|}{-} &
  - &
  \multicolumn{1}{c|}{1678.75} &
  \multicolumn{1}{c|}{1720.40} &
  \multicolumn{1}{c|}{1715.56} &
  \multicolumn{1}{c|}{-} &
  62920.9 \\ \cline{2-12} 
\multicolumn{1}{|c|}{} &
  60 &
  \multicolumn{1}{c|}{2610.41} &
  \multicolumn{1}{c|}{2678.94} &
  \multicolumn{1}{c|}{2655.26} &
  \multicolumn{1}{c|}{-} &
  - &
  \multicolumn{1}{c|}{2708.27} &
  \multicolumn{1}{c|}{2766.59} &
  \multicolumn{1}{c|}{2755.63} &
  \multicolumn{1}{c|}{-} &
  - \\ \cline{2-12} 
\multicolumn{1}{|c|}{} &
  80 &
  \multicolumn{1}{c|}{3616.08} &
  \multicolumn{1}{c|}{3723.15} &
  \multicolumn{1}{c|}{3771.42} &
  \multicolumn{1}{c|}{-} &
  - &
  \multicolumn{1}{c|}{3701.48} &
  \multicolumn{1}{c|}{3813.28} &
  \multicolumn{1}{c|}{3801.57} &
  \multicolumn{1}{c|}{-} &
  - \\ \cline{2-12} 
\multicolumn{1}{|c|}{} &
  100 &
  \multicolumn{1}{c|}{4655.82} &
  \multicolumn{1}{c|}{4863.40} &
  \multicolumn{1}{c|}{5015.05} &
  \multicolumn{1}{c|}{-} &
  - &
  \multicolumn{1}{c|}{-} &
  \multicolumn{1}{c|}{5081.09} &
  \multicolumn{1}{c|}{5158.74} &
  \multicolumn{1}{c|}{-} &
  - \\ \cline{2-12} 
\multicolumn{1}{|c|}{} &
  120 &
  \multicolumn{1}{c|}{-} &
  \multicolumn{1}{c|}{6327.18} &
  \multicolumn{1}{c|}{6344.44} &
  \multicolumn{1}{c|}{-} &
  - &
  \multicolumn{1}{c|}{-} &
  \multicolumn{1}{c|}{6674.42} &
  \multicolumn{1}{c|}{10218.04} &
  \multicolumn{1}{c|}{-} &
  - \\ \cline{2-12} 
\multicolumn{1}{|c|}{} &
  140 &
  \multicolumn{1}{c|}{-} &
  \multicolumn{1}{c|}{7768.68} &
  \multicolumn{1}{c|}{8244.31} &
  \multicolumn{1}{c|}{-} &
  - &
  \multicolumn{1}{c|}{-} &
  \multicolumn{1}{c|}{8405.47} &
  \multicolumn{1}{c|}{12642.75} &
  \multicolumn{1}{c|}{-} &
  - \\ \cline{2-12} 
\multicolumn{1}{|c|}{} &
  160 &
  \multicolumn{1}{c|}{-} &
  \multicolumn{1}{c|}{9622.02} &
  \multicolumn{1}{c|}{11506.63} &
  \multicolumn{1}{c|}{-} &
  - &
  \multicolumn{1}{c|}{-} &
  \multicolumn{1}{c|}{10297.65} &
  \multicolumn{1}{c|}{13091.19} &
  \multicolumn{1}{c|}{-} &
  - \\ \hline
\multicolumn{1}{|c|}{\multirow{8}{*}{Sum of distance (m)}} &
  20 &
  \multicolumn{1}{c|}{397.03} &
  \multicolumn{1}{c|}{399.89} &
  \multicolumn{1}{c|}{399.20} &
  \multicolumn{1}{c|}{458.2} &
  488.9 &
  \multicolumn{1}{c|}{402.88} &
  \multicolumn{1}{c|}{407.30} &
  \multicolumn{1}{c|}{405.27} &
  \multicolumn{1}{c|}{482.3} &
  509.1 \\ \cline{2-12} 
\multicolumn{1}{|c|}{} &
  40 &
  \multicolumn{1}{c|}{793.14} &
  \multicolumn{1}{c|}{802.56} &
  \multicolumn{1}{c|}{801.70} &
  \multicolumn{1}{c|}{-} &
  - &
  \multicolumn{1}{c|}{800.53} &
  \multicolumn{1}{c|}{812.18} &
  \multicolumn{1}{c|}{810.99} &
  \multicolumn{1}{c|}{-} &
  953.2 \\ \cline{2-12} 
\multicolumn{1}{|c|}{} &
  60 &
  \multicolumn{1}{c|}{1218.09} &
  \multicolumn{1}{c|}{1246.38} &
  \multicolumn{1}{c|}{1239.95} &
  \multicolumn{1}{c|}{-} &
  - &
  \multicolumn{1}{c|}{1255.49} &
  \multicolumn{1}{c|}{1281.13} &
  \multicolumn{1}{c|}{1278.15} &
  \multicolumn{1}{c|}{-} &
  - \\ \cline{2-12} 
\multicolumn{1}{|c|}{} &
  80 &
  \multicolumn{1}{c|}{1659.89} &
  \multicolumn{1}{c|}{1707.72} &
  \multicolumn{1}{c|}{1700.10} &
  \multicolumn{1}{c|}{-} &
  - &
  \multicolumn{1}{c|}{1681.96} &
  \multicolumn{1}{c|}{1739.45} &
  \multicolumn{1}{c|}{1728.62} &
  \multicolumn{1}{c|}{-} &
  - \\ \cline{2-12} 
\multicolumn{1}{|c|}{} &
  100 &
  \multicolumn{1}{c|}{2111.40} &
  \multicolumn{1}{c|}{2211.05} &
  \multicolumn{1}{c|}{2213.59} &
  \multicolumn{1}{c|}{-} &
  - &
  \multicolumn{1}{c|}{-} &
  \multicolumn{1}{c|}{2273.16} &
  \multicolumn{1}{c|}{2257.23} &
  \multicolumn{1}{c|}{-} &
  - \\ \cline{2-12} 
\multicolumn{1}{|c|}{} &
  120 &
  \multicolumn{1}{c|}{-} &
  \multicolumn{1}{c|}{2807.80} &
  \multicolumn{1}{c|}{2802.05} &
  \multicolumn{1}{c|}{-} &
  - &
  \multicolumn{1}{c|}{-} &
  \multicolumn{1}{c|}{2898.26} &
  \multicolumn{1}{c|}{2952.09} &
  \multicolumn{1}{c|}{-} &
  - \\ \cline{2-12} 
\multicolumn{1}{|c|}{} &
  140 &
  \multicolumn{1}{c|}{-} &
  \multicolumn{1}{c|}{3372.91} &
  \multicolumn{1}{c|}{3418.93} &
  \multicolumn{1}{c|}{-} &
  - &
  \multicolumn{1}{c|}{-} &
  \multicolumn{1}{c|}{3547.59} &
  \multicolumn{1}{c|}{3628.51} &
  \multicolumn{1}{c|}{-} &
  - \\ \cline{2-12} 
\multicolumn{1}{|c|}{} &
  160 &
  \multicolumn{1}{c|}{-} &
  \multicolumn{1}{c|}{4070.74} &
  \multicolumn{1}{c|}{4155.97} &
  \multicolumn{1}{c|}{-} &
  - &
  \multicolumn{1}{c|}{-} &
  \multicolumn{1}{c|}{4230.00} &
  \multicolumn{1}{c|}{4422.56} &
  \multicolumn{1}{c|}{-} &
  - \\ \hline
\multicolumn{1}{|c|}{\multirow{8}{*}{Makespan (sec)}} &
  20 &
  \multicolumn{1}{c|}{73.34} &
  \multicolumn{1}{c|}{73.82} &
  \multicolumn{1}{c|}{73.56} &
  \multicolumn{1}{c|}{106.9} &
  1243.3 &
  \multicolumn{1}{c|}{75.41} &
  \multicolumn{1}{c|}{75.91} &
  \multicolumn{1}{c|}{75.61} &
  \multicolumn{1}{c|}{112.0} &
  1283.3 \\ \cline{2-12} 
\multicolumn{1}{|c|}{} &
  40 &
  \multicolumn{1}{c|}{77.87} &
  \multicolumn{1}{c|}{79.08} &
  \multicolumn{1}{c|}{78.65} &
  \multicolumn{1}{c|}{-} &
  - &
  \multicolumn{1}{c|}{82.85} &
  \multicolumn{1}{c|}{84.20} &
  \multicolumn{1}{c|}{86.55} &
  \multicolumn{1}{c|}{-} &
  2387.0 \\ \cline{2-12} 
\multicolumn{1}{|c|}{} &
  60 &
  \multicolumn{1}{c|}{82.60} &
  \multicolumn{1}{c|}{86.08} &
  \multicolumn{1}{c|}{86.45} &
  \multicolumn{1}{c|}{-} &
  - &
  \multicolumn{1}{c|}{84.55} &
  \multicolumn{1}{c|}{88.11} &
  \multicolumn{1}{c|}{90.68} &
  \multicolumn{1}{c|}{-} &
  - \\ \cline{2-12} 
\multicolumn{1}{|c|}{} &
  80 &
  \multicolumn{1}{c|}{84.67} &
  \multicolumn{1}{c|}{88.92} &
  \multicolumn{1}{c|}{111.96} &
  \multicolumn{1}{c|}{-} &
  - &
  \multicolumn{1}{c|}{85.23} &
  \multicolumn{1}{c|}{95.89} &
  \multicolumn{1}{c|}{102.26} &
  \multicolumn{1}{c|}{-} &
  - \\ \cline{2-12} 
\multicolumn{1}{|c|}{} &
  100 &
  \multicolumn{1}{c|}{77.61} &
  \multicolumn{1}{c|}{98.03} &
  \multicolumn{1}{c|}{226.64} &
  \multicolumn{1}{c|}{-} &
  - &
  \multicolumn{1}{c|}{-} &
  \multicolumn{1}{c|}{111.54} &
  \multicolumn{1}{c|}{201.24} &
  \multicolumn{1}{c|}{-} &
  - \\ \cline{2-12} 
\multicolumn{1}{|c|}{} &
  120 &
  \multicolumn{1}{c|}{-} &
  \multicolumn{1}{c|}{110.11} &
  \multicolumn{1}{c|}{125.12} &
  \multicolumn{1}{c|}{-} &
  - &
  \multicolumn{1}{c|}{-} &
  \multicolumn{1}{c|}{137.06} &
  \multicolumn{1}{c|}{1775.00} &
  \multicolumn{1}{c|}{-} &
  - \\ \cline{2-12} 
\multicolumn{1}{|c|}{} &
  140 &
  \multicolumn{1}{c|}{-} &
  \multicolumn{1}{c|}{133.60} &
  \multicolumn{1}{c|}{345.70} &
  \multicolumn{1}{c|}{-} &
  - &
  \multicolumn{1}{c|}{-} &
  \multicolumn{1}{c|}{151.38} &
  \multicolumn{1}{c|}{1329.46} &
  \multicolumn{1}{c|}{-} &
  - \\ \cline{2-12} 
\multicolumn{1}{|c|}{} &
  160 &
  \multicolumn{1}{c|}{-} &
  \multicolumn{1}{c|}{153.98} &
  \multicolumn{1}{c|}{1227.89} &
  \multicolumn{1}{c|}{-} &
  - &
  \multicolumn{1}{c|}{-} &
  \multicolumn{1}{c|}{165.12} &
  \multicolumn{1}{c|}{1344.00} &
  \multicolumn{1}{c|}{-} &
  - \\ \hline
\end{tabular}
\end{subtable}%
}\vspace{5pt}
\resizebox{\columnwidth}{!}{%
\begin{subtable}[t]{\textwidth}
\centering
\captionsetup{skip=1pt}
\caption{More difficult environments \textsf{Circ20} and \textsf{Rect20}}
\begin{tabular}{|cc|ccccc|ccccc|}
\hline
\multicolumn{2}{|c|}{Environment} &
  \multicolumn{5}{c|}{\textsf{Circ20}} &
  \multicolumn{5}{c|}{\textsf{Rect20}} \\ \hline
\multicolumn{1}{|c|}{Metric} &
  \#robot &
  \multicolumn{1}{c|}{SI-CCBS} &
  \multicolumn{1}{c|}{SI-CPP} &
  \multicolumn{1}{c|}{\strrtpp} &
  \multicolumn{1}{c|}{GT} &
  SSSP &
  \multicolumn{1}{c|}{SI-CCBS} &
  \multicolumn{1}{c|}{SI-CPP} &
  \multicolumn{1}{c|}{\strrtpp} &
  \multicolumn{1}{c|}{GT} &
  SSSP \\ \hline
\multicolumn{1}{|c|}{\multirow{8}{*}{Success rate (\%)}} &
  20 &
  \multicolumn{1}{c|}{100} &
  \multicolumn{1}{c|}{100} &
  \multicolumn{1}{c|}{100} &
  \multicolumn{1}{c|}{28} &
  34 &
  \multicolumn{1}{c|}{100} &
  \multicolumn{1}{c|}{100} &
  \multicolumn{1}{c|}{100} &
  \multicolumn{1}{c|}{20} &
  22 \\ \cline{2-12} 
\multicolumn{1}{|c|}{} &
  40 &
  \multicolumn{1}{c|}{100} &
  \multicolumn{1}{c|}{100} &
  \multicolumn{1}{c|}{100} &
  \multicolumn{1}{c|}{0} &
  0 &
  \multicolumn{1}{c|}{100} &
  \multicolumn{1}{c|}{100} &
  \multicolumn{1}{c|}{100} &
  \multicolumn{1}{c|}{0} &
  0 \\ \cline{2-12} 
\multicolumn{1}{|c|}{} &
  60 &
  \multicolumn{1}{c|}{92} &
  \multicolumn{1}{c|}{100} &
  \multicolumn{1}{c|}{100} &
  \multicolumn{1}{c|}{0} &
  0 &
  \multicolumn{1}{c|}{92} &
  \multicolumn{1}{c|}{100} &
  \multicolumn{1}{c|}{100} &
  \multicolumn{1}{c|}{0} &
  0 \\ \cline{2-12} 
\multicolumn{1}{|c|}{} &
  80 &
  \multicolumn{1}{c|}{38} &
  \multicolumn{1}{c|}{100} &
  \multicolumn{1}{c|}{96} &
  \multicolumn{1}{c|}{0} &
  0 &
  \multicolumn{1}{c|}{20} &
  \multicolumn{1}{c|}{98} &
  \multicolumn{1}{c|}{96} &
  \multicolumn{1}{c|}{0} &
  0 \\ \cline{2-12} 
\multicolumn{1}{|c|}{} &
  100 &
  \multicolumn{1}{c|}{2} &
  \multicolumn{1}{c|}{100} &
  \multicolumn{1}{c|}{98} &
  \multicolumn{1}{c|}{0} &
  0 &
  \multicolumn{1}{c|}{0} &
  \multicolumn{1}{c|}{100} &
  \multicolumn{1}{c|}{86} &
  \multicolumn{1}{c|}{0} &
  0 \\ \cline{2-12} 
\multicolumn{1}{|c|}{} &
  120 &
  \multicolumn{1}{c|}{0} &
  \multicolumn{1}{c|}{100} &
  \multicolumn{1}{c|}{82} &
  \multicolumn{1}{c|}{0} &
  0 &
  \multicolumn{1}{c|}{0} &
  \multicolumn{1}{c|}{96} &
  \multicolumn{1}{c|}{62} &
  \multicolumn{1}{c|}{0} &
  0 \\ \cline{2-12} 
\multicolumn{1}{|c|}{} &
  140 &
  \multicolumn{1}{c|}{0} &
  \multicolumn{1}{c|}{96} &
  \multicolumn{1}{c|}{66} &
  \multicolumn{1}{c|}{0} &
  0 &
  \multicolumn{1}{c|}{0} &
  \multicolumn{1}{c|}{96} &
  \multicolumn{1}{c|}{44} &
  \multicolumn{1}{c|}{0} &
  0 \\ \cline{2-12} 
\multicolumn{1}{|c|}{} &
  160 &
  \multicolumn{1}{c|}{0} &
  \multicolumn{1}{c|}{98} &
  \multicolumn{1}{c|}{18} &
  \multicolumn{1}{c|}{0} &
  0 &
  \multicolumn{1}{c|}{0} &
  \multicolumn{1}{c|}{62} &
  \multicolumn{1}{c|}{12} &
  \multicolumn{1}{c|}{0} &
  0 \\ \hline
\multicolumn{1}{|c|}{\multirow{8}{*}{Flowtime (sec)}} &
  20 &
  \multicolumn{1}{c|}{837.47} &
  \multicolumn{1}{c|}{857.39} &
  \multicolumn{1}{c|}{873.1} &
  \multicolumn{1}{c|}{2295.3} &
  17142.9 &
  \multicolumn{1}{c|}{850.38} &
  \multicolumn{1}{c|}{870.72} &
  \multicolumn{1}{c|}{866.1} &
  \multicolumn{1}{c|}{2364.9} &
  16696.4 \\ \cline{2-12} 
\multicolumn{1}{|c|}{} &
  40 &
  \multicolumn{1}{c|}{1791.15} &
  \multicolumn{1}{c|}{1848.30} &
  \multicolumn{1}{c|}{1854.0} &
  \multicolumn{1}{c|}{-} &
  - &
  \multicolumn{1}{c|}{1839.61} &
  \multicolumn{1}{c|}{1893.12} &
  \multicolumn{1}{c|}{1910.0} &
  \multicolumn{1}{c|}{-} &
  - \\ \cline{2-12} 
\multicolumn{1}{|c|}{} &
  60 &
  \multicolumn{1}{c|}{2800.59} &
  \multicolumn{1}{c|}{2907.90} &
  \multicolumn{1}{c|}{2994.7} &
  \multicolumn{1}{c|}{-} &
  - &
  \multicolumn{1}{c|}{2955.70} &
  \multicolumn{1}{c|}{3129.13} &
  \multicolumn{1}{c|}{3763.5} &
  \multicolumn{1}{c|}{-} &
  - \\ \cline{2-12} 
\multicolumn{1}{|c|}{} &
  80 &
  \multicolumn{1}{c|}{3911.63} &
  \multicolumn{1}{c|}{4193.95} &
  \multicolumn{1}{c|}{5052.9} &
  \multicolumn{1}{c|}{-} &
  - &
  \multicolumn{1}{c|}{3924.30} &
  \multicolumn{1}{c|}{4581.84} &
  \multicolumn{1}{c|}{5093.6} &
  \multicolumn{1}{c|}{-} &
  - \\ \cline{2-12} 
\multicolumn{1}{|c|}{} &
  100 &
  \multicolumn{1}{c|}{4253.73} &
  \multicolumn{1}{c|}{5685.72} &
  \multicolumn{1}{c|}{6307.6} &
  \multicolumn{1}{c|}{-} &
  - &
  \multicolumn{1}{c|}{-} &
  \multicolumn{1}{c|}{6341.40} &
  \multicolumn{1}{c|}{10115.3} &
  \multicolumn{1}{c|}{-} &
  - \\ \cline{2-12} 
\multicolumn{1}{|c|}{} &
  120 &
  \multicolumn{1}{c|}{-} &
  \multicolumn{1}{c|}{7585.49} &
  \multicolumn{1}{c|}{10901.93} &
  \multicolumn{1}{c|}{-} &
  - &
  \multicolumn{1}{c|}{-} &
  \multicolumn{1}{c|}{8512.18} &
  \multicolumn{1}{c|}{14308.43} &
  \multicolumn{1}{c|}{-} &
  - \\ \cline{2-12} 
\multicolumn{1}{|c|}{} &
  140 &
  \multicolumn{1}{c|}{-} &
  \multicolumn{1}{c|}{9599.20} &
  \multicolumn{1}{c|}{12278.50} &
  \multicolumn{1}{c|}{-} &
  - &
  \multicolumn{1}{c|}{-} &
  \multicolumn{1}{c|}{10650.45} &
  \multicolumn{1}{c|}{14727.81} &
  \multicolumn{1}{c|}{-} &
  - \\ \cline{2-12} 
\multicolumn{1}{|c|}{} &
  160 &
  \multicolumn{1}{c|}{-} &
  \multicolumn{1}{c|}{11942.79} &
  \multicolumn{1}{c|}{16263.49} &
  \multicolumn{1}{c|}{-} &
  - &
  \multicolumn{1}{c|}{-} &
  \multicolumn{1}{c|}{13455.49} &
  \multicolumn{1}{c|}{20182.43} &
  \multicolumn{1}{c|}{-} &
  - \\ \hline
\multicolumn{1}{|c|}{\multirow{8}{*}{Sum of distance (m)}} &
  20 &
  \multicolumn{1}{c|}{410.34} &
  \multicolumn{1}{c|}{416.08} &
  \multicolumn{1}{c|}{417.6} &
  \multicolumn{1}{c|}{443.5} &
  507.9 &
  \multicolumn{1}{c|}{410.11} &
  \multicolumn{1}{c|}{418.53} &
  \multicolumn{1}{c|}{413.6} &
  \multicolumn{1}{c|}{468.6} &
  507.3 \\ \cline{2-12} 
\multicolumn{1}{|c|}{} &
  40 &
  \multicolumn{1}{c|}{849.90} &
  \multicolumn{1}{c|}{870.19} &
  \multicolumn{1}{c|}{867.0} &
  \multicolumn{1}{c|}{-} &
  - &
  \multicolumn{1}{c|}{858.95} &
  \multicolumn{1}{c|}{880.18} &
  \multicolumn{1}{c|}{876.5} &
  \multicolumn{1}{c|}{-} &
  - \\ \cline{2-12} 
\multicolumn{1}{|c|}{} &
  60 &
  \multicolumn{1}{c|}{1287.37} &
  \multicolumn{1}{c|}{1335.40} &
  \multicolumn{1}{c|}{1350.0} &
  \multicolumn{1}{c|}{-} &
  - &
  \multicolumn{1}{c|}{1325.91} &
  \multicolumn{1}{c|}{1404.70} &
  \multicolumn{1}{c|}{1400.8} &
  \multicolumn{1}{c|}{-} &
  - \\ \cline{2-12} 
\multicolumn{1}{|c|}{} &
  80 &
  \multicolumn{1}{c|}{1758.10} &
  \multicolumn{1}{c|}{1869.69} &
  \multicolumn{1}{c|}{1899.5} &
  \multicolumn{1}{c|}{-} &
  - &
  \multicolumn{1}{c|}{1714.89} &
  \multicolumn{1}{c|}{1968.39} &
  \multicolumn{1}{c|}{1978.7} &
  \multicolumn{1}{c|}{-} &
  - \\ \cline{2-12} 
\multicolumn{1}{|c|}{} &
  100 &
  \multicolumn{1}{c|}{1972.24} &
  \multicolumn{1}{c|}{2448.97} &
  \multicolumn{1}{c|}{2532.8} &
  \multicolumn{1}{c|}{-} &
  - &
  \multicolumn{1}{c|}{-} &
  \multicolumn{1}{c|}{2629.45} &
  \multicolumn{1}{c|}{2675.4} &
  \multicolumn{1}{c|}{-} &
  - \\ \cline{2-12} 
\multicolumn{1}{|c|}{} &
  120 &
  \multicolumn{1}{c|}{-} &
  \multicolumn{1}{c|}{3176.30} &
  \multicolumn{1}{c|}{3328.48} &
  \multicolumn{1}{c|}{-} &
  - &
  \multicolumn{1}{c|}{-} &
  \multicolumn{1}{c|}{3375.16} &
  \multicolumn{1}{c|}{3688.02} &
  \multicolumn{1}{c|}{-} &
  - \\ \cline{2-12} 
\multicolumn{1}{|c|}{} &
  140 &
  \multicolumn{1}{c|}{-} &
  \multicolumn{1}{c|}{3871.72} &
  \multicolumn{1}{c|}{4093.30} &
  \multicolumn{1}{c|}{-} &
  - &
  \multicolumn{1}{c|}{-} &
  \multicolumn{1}{c|}{4090.95} &
  \multicolumn{1}{c|}{4589.90} &
  \multicolumn{1}{c|}{-} &
  - \\ \cline{2-12} 
\multicolumn{1}{|c|}{} &
  160 &
  \multicolumn{1}{c|}{-} &
  \multicolumn{1}{c|}{4614.01} &
  \multicolumn{1}{c|}{5201.03} &
  \multicolumn{1}{c|}{-} &
  - &
  \multicolumn{1}{c|}{-} &
  \multicolumn{1}{c|}{4944.88} &
  \multicolumn{1}{c|}{5598.63} &
  \multicolumn{1}{c|}{-} &
  - \\ \hline
\multicolumn{1}{|c|}{\multirow{8}{*}{Makespan (sec)}} &
  20 &
  \multicolumn{1}{c|}{77.27} &
  \multicolumn{1}{c|}{78.03} &
  \multicolumn{1}{c|}{83.4} &
  \multicolumn{1}{c|}{114.8} &
  1286.1 &
  \multicolumn{1}{c|}{79.88} &
  \multicolumn{1}{c|}{82.14} &
  \multicolumn{1}{c|}{82.4} &
  \multicolumn{1}{c|}{118.2} &
  1259.6 \\ \cline{2-12} 
\multicolumn{1}{|c|}{} &
  40 &
  \multicolumn{1}{c|}{84.49} &
  \multicolumn{1}{c|}{88.75} &
  \multicolumn{1}{c|}{92.6} &
  \multicolumn{1}{c|}{-} &
  - &
  \multicolumn{1}{c|}{86.36} &
  \multicolumn{1}{c|}{91.58} &
  \multicolumn{1}{c|}{98.1} &
  \multicolumn{1}{c|}{-} &
  - \\ \cline{2-12} 
\multicolumn{1}{|c|}{} &
  60 &
  \multicolumn{1}{c|}{86.81} &
  \multicolumn{1}{c|}{97.16} &
  \multicolumn{1}{c|}{114.1} &
  \multicolumn{1}{c|}{-} &
  - &
  \multicolumn{1}{c|}{87.57} &
  \multicolumn{1}{c|}{117.29} &
  \multicolumn{1}{c|}{458.8} &
  \multicolumn{1}{c|}{-} &
  - \\ \cline{2-12} 
\multicolumn{1}{|c|}{} &
  80 &
  \multicolumn{1}{c|}{86.64} &
  \multicolumn{1}{c|}{111.95} &
  \multicolumn{1}{c|}{514.2} &
  \multicolumn{1}{c|}{-} &
  - &
  \multicolumn{1}{c|}{89.80} &
  \multicolumn{1}{c|}{132.38} &
  \multicolumn{1}{c|}{502.7} &
  \multicolumn{1}{c|}{-} &
  - \\ \cline{2-12} 
\multicolumn{1}{|c|}{} &
  100 &
  \multicolumn{1}{c|}{74.25} &
  \multicolumn{1}{c|}{138.00} &
  \multicolumn{1}{c|}{294.5} &
  \multicolumn{1}{c|}{-} &
  - &
  \multicolumn{1}{c|}{-} &
  \multicolumn{1}{c|}{160.33} &
  \multicolumn{1}{c|}{1769.9} &
  \multicolumn{1}{c|}{-} &
  - \\ \cline{2-12} 
\multicolumn{1}{|c|}{} &
  120 &
  \multicolumn{1}{c|}{-} &
  \multicolumn{1}{c|}{155.34} &
  \multicolumn{1}{c|}{1973.52} &
  \multicolumn{1}{c|}{-} &
  - &
  \multicolumn{1}{c|}{-} &
  \multicolumn{1}{c|}{176.82} &
  \multicolumn{1}{c|}{2084.40} &
  \multicolumn{1}{c|}{-} &
  - \\ \cline{2-12} 
\multicolumn{1}{|c|}{} &
  140 &
  \multicolumn{1}{c|}{-} &
  \multicolumn{1}{c|}{173.34} &
  \multicolumn{1}{c|}{613.73} &
  \multicolumn{1}{c|}{-} &
  - &
  \multicolumn{1}{c|}{-} &
  \multicolumn{1}{c|}{182.75} &
  \multicolumn{1}{c|}{587.23} &
  \multicolumn{1}{c|}{-} &
  - \\ \cline{2-12} 
\multicolumn{1}{|c|}{} &
  160 &
  \multicolumn{1}{c|}{-} &
  \multicolumn{1}{c|}{189.84} &
  \multicolumn{1}{c|}{530.44} &
  \multicolumn{1}{c|}{-} &
  - &
  \multicolumn{1}{c|}{-} &
  \multicolumn{1}{c|}{206.43} &
  \multicolumn{1}{c|}{880.44} &
  \multicolumn{1}{c|}{-} &
  - \\ \hline
\end{tabular}
\end{subtable}%
}
\end{table}

\section{Dynamic simulation}
\begin{figure}[h]
    \centering    
    \begin{subfigure}{\textwidth}
    \centering    
        \includegraphics[width=0.2\linewidth]{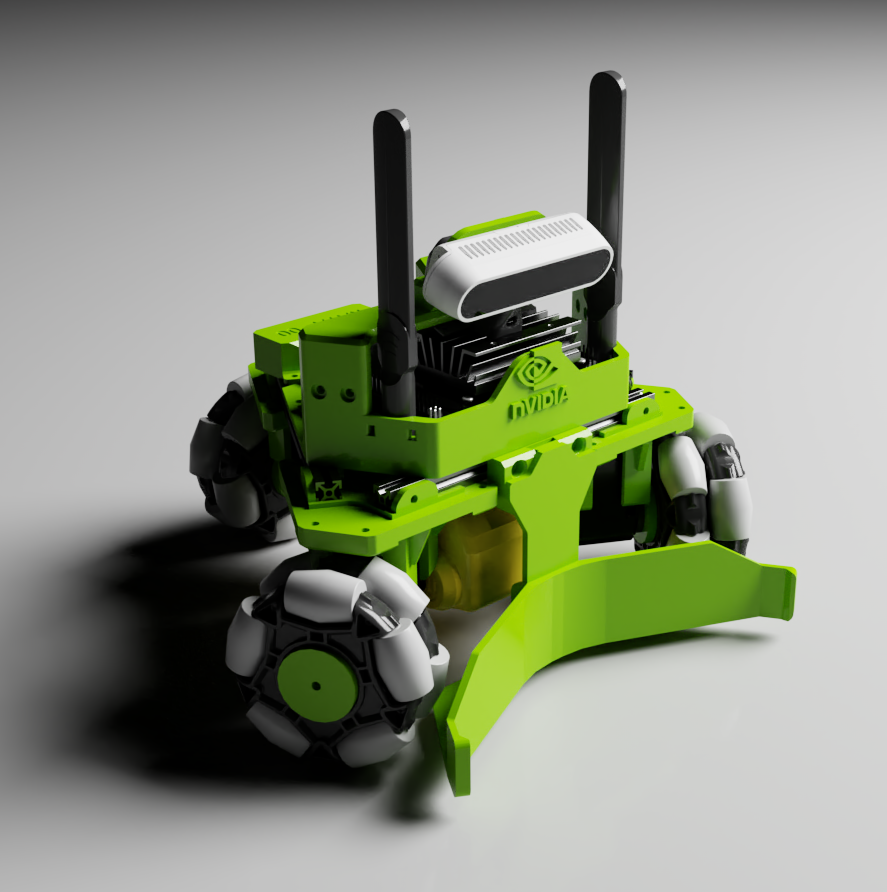}
        \caption{}
        \label{fig:robot}
    \end{subfigure}
    \begin{subfigure}{\textwidth}
    \centering    
        \includegraphics[width=0.9\linewidth]{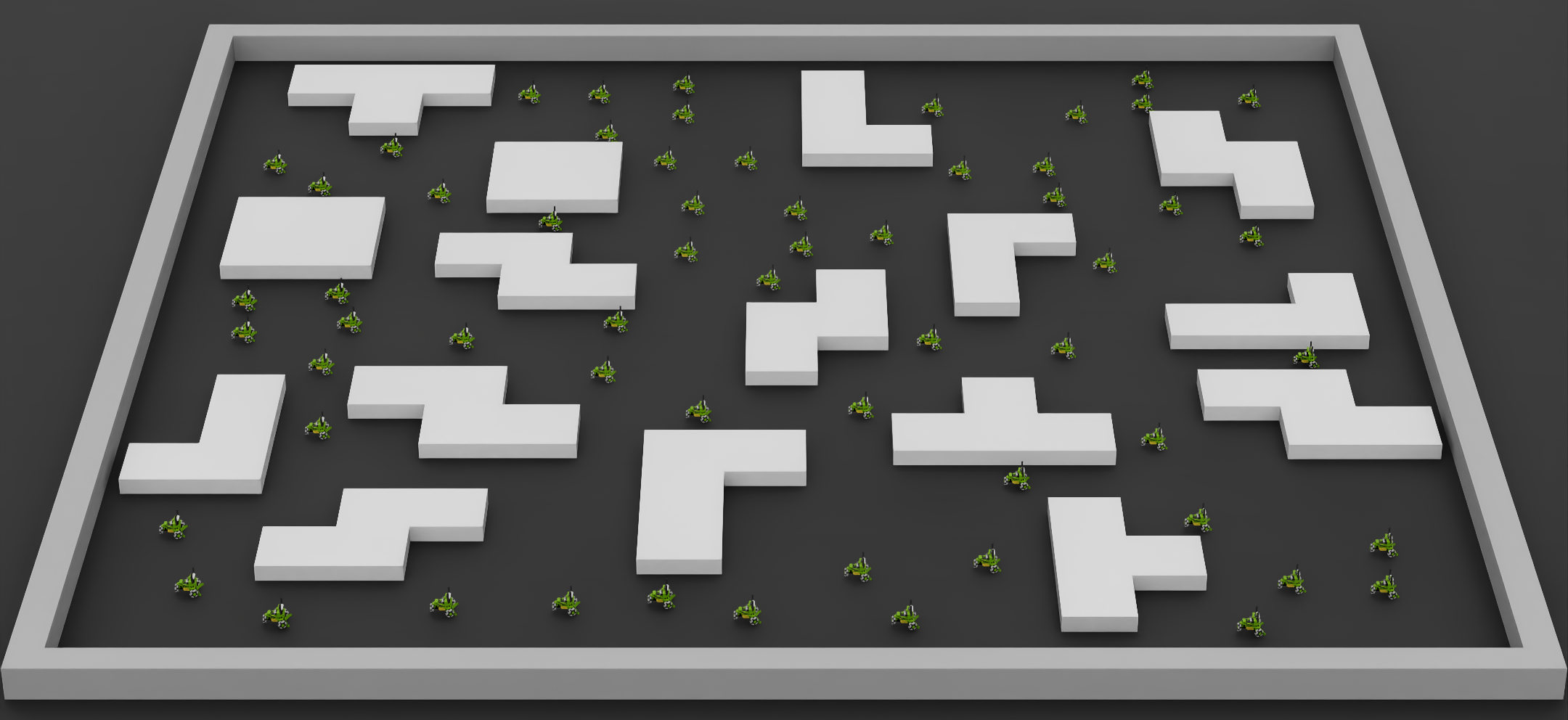}
        \caption{}
        \label{fig:env}
    \end{subfigure}
    \caption{ (a) Kaya robot (omnidirectional) (b) A tested instance of \textsf{Rect20} with $60$ robots in NVIDIA Isaac Sim}
    \label{fig:dynamic}
\end{figure}

Since \sirrt assumes that robots move with a fixed velocity, the experimental results might be less convincing to use our methods for physical robots with complex dynamics. Thus, we executed our solutions from \sipp by robots in NVIDIA Isaac Sim (\url{https://developer.nvidia.com/isaac-sim}), a dynamic simulator with a high-fidelity physics engine. Since loading tens of robots in Isaac Sim is resource-intensive, we chose one of the simple robots, Kaya shown in Fig.~\ref{fig:robot}. We provide a safety margin of 20\% of the radius of the robot for each robot, a reasonable setup in real-robot implementations. Given the computational resource of a PC with the exact specification in the main manuscript, we could reliably run $60$ robots in Isaac Sim. 

A solution of \sipp provides acceleration values to control velocity, along with the durations for acceleration and deceleration. The robots reach their destinations by accelerating and decelerating at the specified rates and times according to the given control inputs. Specifically, they accelerate at the determined rate for the specified duration and then decelerate at the determined rate for the specified duration. A solution is valid if no collision or deadlock occurs during the navigation. Among $20$ test instances in the most challenging environment \textsf{Rect20} (Fig.~\ref{fig:env}), $100\%$ of the solutions were turned out to be valid. The supplementary video includes two example executions in \textsf{Circ20} and \textsf{Rect20}. 

Overall, the execution result of solutions in dynamic simulation shows the feasibility of the proposed methods in real-world applications.

